\documentclass[10pt,twocolumn,letterpaper]{article}

\usepackage[pagenumbers]{wacv} 

\usepackage{multirow}
\usepackage{adjustbox}
\usepackage{graphicx}

\usepackage[table]{xcolor}
\definecolor{tabfirst}{rgb}{1, 0.7, 0.7} 
\definecolor{tabsecond}{rgb}{1, 0.85, 0.7} 
\definecolor{tabthird}{rgb}{1, 1, 0.7} 

\definecolor{wacvblue}{rgb}{0.21,0.49,0.74}
\usepackage[pagebackref,breaklinks,colorlinks,allcolors=wacvblue]{hyperref}

\def\wacvPaperID{890} 
\def\confName{WACV}
\def\confYear{2026}

\title{MoRe: Monocular Geometry Refinement via Graph Optimization for Cross-View Consistency}

\author{
Dongki Jung\textsuperscript{\rm 1} \:\: 
Jaehoon Choi\textsuperscript{\rm 1} \:\: 
Yonghan Lee\textsuperscript{\rm 1} \:\: 
Sungmin Eum\textsuperscript{\rm 2} \\
Heesung Kwon\textsuperscript{\rm 2} \:\:
Dinesh Manocha\textsuperscript{\rm 1} \\
$^{1}$University of Maryland, College Park \:\: $^{2}$DEVCOM Army Research Laboratory\\
\vspace*{-4mm}
}

\begin{document}
\maketitle
\begin{abstract}
Monocular 3D foundation models offer an extensible solution for perception tasks, making them attractive for broader 3D vision applications.
In this paper, we propose MoRe, a training-free \underline{Mo}nocular Geometry \underline{Re}finement method designed to improve cross-view consistency and achieve scale alignment.
To induce inter-frame relationships, our method employs feature matching between frames to establish correspondences.
Rather than applying simple least squares optimization on these matched points, we formulate a graph-based optimization framework that performs local planar approximation using the estimated 3D points and surface normals estimated by monocular foundation models.
This formulation addresses the scale ambiguity inherent in monocular geometric priors while preserving the underlying 3D structure.
We further demonstrate that MoRe not only enhances 3D reconstruction but also improves novel view synthesis, particularly in sparse-view rendering scenarios.
\end{abstract}
    
\section{Introduction}
\label{sec:intro}
\begin{figure}[t]
    \centering
    \includegraphics[width=1.\linewidth]{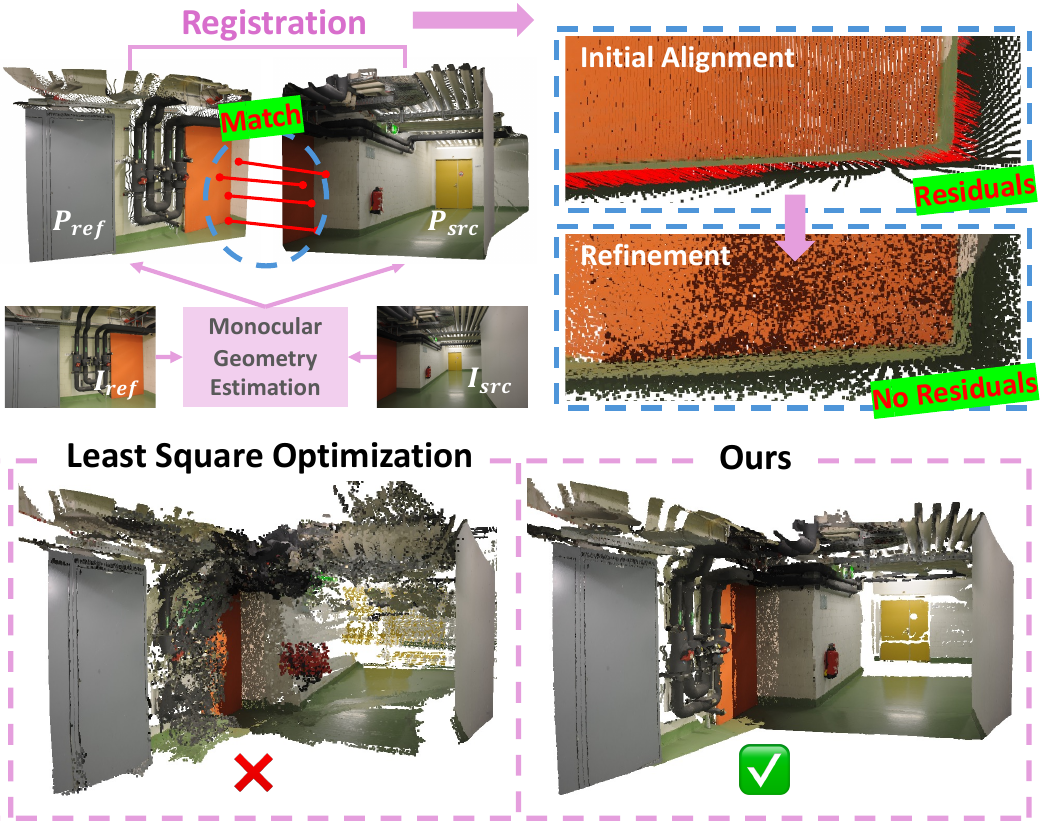}
    \caption{
    Monocular geometry estimation often suffers from scale ambiguity across different views, leading to 3D points with inconsistent scales.
    To address this, we propose MoRe, a monocular geometry refinement method for aligning point maps across views.
    We first apply an initial affine transformation using matched 3D points, followed by a novel refinement step. 
    Red lines indicate residual distances between corresponding points.
    During refinement, instead of directly minimizing least squares error over these correspondences, we introduce a graph-based optimization that yields more accurate and consistent 3D reconstructions.
    }
    \label{fig:toy}
\end{figure}
Foundation models have recently achieved significant progress in 3D vision tasks \cite{hu2024metric3d,wang2024moge,wang2024dust3r,leroy2024grounding,zhang2024monst3r,wang2025vggt}, demonstrating promising results in large-scale 3D reconstruction and scene understanding.
These models benefit from data-driven learning, offering scalable and accessible solutions across a wide range of 3D vision tasks, including Structure-from-Motion (SfM) and SLAM \cite{pataki2025mp, yu2025relative, duisterhof2024mast3r, murai2024mast3r, elflein2025light3r}.
Recent 3D foundation models have shown that point maps can implicitly capture relationships between pixels and the underlying 3D scene \cite{wang2024moge, wang2024dust3r}.
However, since the point maps estimated from each camera do not share a common coordinate system, they often suffer from misalignments caused by scale ambiguities.
To mitigate this issue, recent works \cite{wang2024dust3r, leroy2024grounding, wang2025vggt} have adopted end-to-end multi-view settings or proposed global alignment methods to enforce a shared coordinate system across views.
While these approaches improve cross-view consistency, they still exhibit several limitations.
First, the lack of modularity in end-to-end frameworks makes it difficult to incorporate additional sensors, such as LiDAR, which are critical for applications in robotics and AR/VR.
Incorporating new sensors typically requires retraining the entire model with sufficiently large datasets, posing significant challenges for adaptability.
Furthermore, due to their reliance on latent feature spaces, these models do not guarantee compatibility with explicit geometric constraints or external sensor modalities.
Another challenge is that multi-view foundation models often operate within their own internal coordinate systems.
Since the training objective primarily focuses on pointmap estimation \cite{wang2024dust3r}, the predicted camera poses are aligned with the estimated points rather than real-world coordinates.
As a result, achieving accurate visual localization or global alignment becomes inherently difficult.
To overcome these limitations, we propose a novel framework that leverages monocular 3D foundation models \cite{wang2024moge} as modular components to achieve consistent 3D reconstruction.
By decoupling point map alignment from full 3D scene reconstruction, our approach enables flexible integration with traditional geometry-based methods, while still benefiting from data-driven pointmap estimation \cite{wang2024dust3r}.
This hybrid formulation combines the strengths of data-driven learning and geometric reasoning, offering the way for practical deployment in real-world scenarios that demand both scalability and reliability.

\paragraph{Main Results} 
In this paper, we present MoRe, a novel monocular point map refinement method designed to enhance cross-view consistency.
When poses are available from off-the-shelf algorithms such as Structure-from-Motion, surface normals can be transformed into a common world coordinate system, serving as strong geometric priors for 3D reconstruction.
By leveraging a joint graph-based optimization that incorporates both estimated 3D points and surface normals, our method enables cross-view alignment of point maps derived from monocular 3D foundation models.
To further incorporate inter-view relationships into the optimization process, we employ image matching algorithms such as \cite{edstedt2023dkm} to establish dense correspondences between images from different views.
A straightforward approach to aligning 3D points in different coordinates is to use the matching points and directly minimize the Euclidean distance between them. 
However, as shown in the least squares optimization example in Fig. \ref{fig:toy}, this method is prone to significant noise.
Instead, we adopt a local planar approximation within the graph optimization, enhancing geometric accuracy of surface reconstruction.

\begin{itemize}
    \item We introduce a novel method for explicitly aligning point maps predicted by monocular 3D foundation models across views, enabling consistent 3D representations.
    \item We propose a graph-based optimization method that incorporates cross-view 3D points and surface normals with local planar constraints for geometric alignment.
    \item We demonstrate that our monocular point map alignment improves novel view rendering performance, particularly in sparse-view scenarios.
\end{itemize}
\section{Related Work}
\label{sec:related}

\paragraph{Foundation Models for 3D Reconstruction}
3D reconstruction is a fundamental problem in computer vision, encompassing a range of tasks such as depth estimation\cite{torralba2002depth}, Structure-from-Motion \cite{wu2013towards,schonberger2016structure} and Multi-view Stereo \cite{campbell2008using,furukawa2015multi,schonberger2016pixelwise}.
Following the emergence of deep learning, substantial research efforts have shifted toward using large-scale datasets to train neural networks for various 3D reconstruction tasks. 
Early research focused on monocular depth estimation \cite{eigen2014depth,laina2016deeper,fu2018deep}, driven either by supervised learning with annotated datasets \cite{eigen2014depth,li2018megadepth,li2019learning,ranftl2020towards} or by self-supervised training methods \cite{garg2016unsupervised,godard2017unsupervised,zhou2017unsupervised,godard2019digging,jung2021dnd,choi2022selftune}.
In particular, MiDaS \cite{ranftl2020towards} demonstrates the effectiveness of supervised training through its zero-shot performance in depth estimation. 
Since then, a dominant line of work \cite{yang2024depth,yang2024depthv2,piccinelli2024unidepth,hu2024metric3d} has focused on collecting large scale datasets using both synthetic and real world data to achieve robust performance across diverse scenarios.
Metric3Dv2 \cite{hu2024metric3d} have designed training strategies to estimate both metric depth and surface normals using these large-scale datasets.

More recently, DUSt3R \cite{wang2024dust3r} and MoGe \cite{wang2024moge} introduced the point map representation, demonstrating its potential for improved geometric performance.
DUSt3R proposed an end-to-end model that predicts globally consistent point maps from two views.
VGGT \cite{wang2025vggt} extended this idea to multi-view inference, enabling global pointmap estimation across multiple images.
Building on this success, many concurrent works have explored various domains, including dynamic scenes \cite{zhang2024monst3r}, structure-from-motion \cite{duisterhof2024mast3r,leroy2024grounding,elflein2025light3r,wang2025vggt}.
While DUSt3R achieves strong results in an end-to-end setting, its reconstruction accuracy tends to degrade when relying on externally provided camera poses rather than jointly estimating them.
Although VGGT improves both reconstruction and pose estimation performance, its architecture does not allow the use of externally given poses.
Thus, we propose a new approach that utilizes monocular 3D foundation models and aligns point maps across different views under given camera poses.

\section{Our Approach: MoRe}
\label{sec:method}
%
%
We propose a monocular 3D reconstruction method to capture the structural consistency and geometric alignment between two distinct images, $\textbf{I}^{\text{ref}}$ and $\textbf{I}^{\text{src}}$, by leveraging both intra- and inter-frame relationships.
In Section \ref{section:initial_align}, we employ a monocular foundation model \cite{wang2024moge} to estimate point maps from the two input views. 
These predicted point maps are then initially aligned to ensure view consistency across the two images.
In Section \ref{section:refinement}, we describe a graph-based optimization process that refines the alignment using the estimated point maps. 
This refinement incorporates geometric constraints into the graph design to improve the accuracy and structural coherence of the alignment.

\begin{figure}[t]
    \centering
    \includegraphics[width=1.\linewidth]{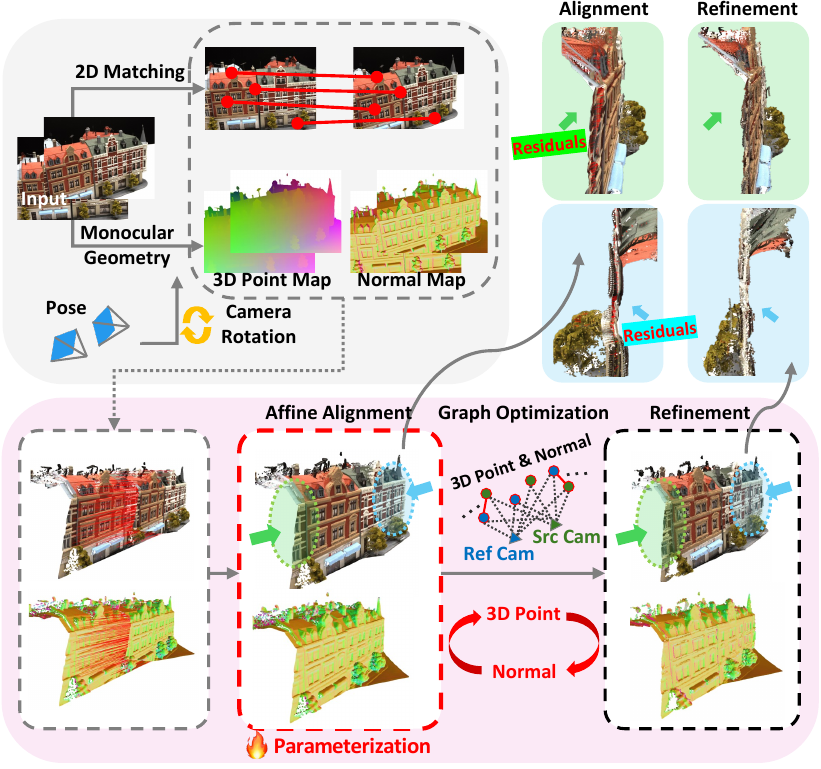}
    \caption{
    Overview of our proposed method.
    Given input images and camera poses, we first generate monocular point maps and surface normal maps using a 3D foundation model.
    We then perform initial alignment using 2D feature correspondences and estimate an affine transformation (scale and shift) to roughly align point maps across views.
    As shown in the Alignment visualization (top right), the initial alignment brings 3D points into a similar position, but residual errors (red lines) still remain.
    To further improve consistency, we introduce a graph-based optimization that jointly parameterizes 3D points and surface normals to refine alignment at the pixel level.
    This refinement significantly reduces residuals and improves geometric coherence across views.
    }
    \label{fig:pipeline}
\end{figure}
\subsection{Scale and Shift Alignment for Point Maps}
\label{section:initial_align}
For an input image $\textbf{I} \in \mathbb{R}^{W \times H \times 3}$ of resolution $W \times H$, the monocular foundation model $F_{\theta}$ \cite{wang2024moge} predicts a 3D point for each pixel, producing a point map $\widehat{\textbf{P}} \in \mathbb{R}^{W \times H \times 3}$.
To establish the geometric relationship between two images $\textbf{I}^{\text{ref}}$ and $\textbf{I}^{\text{src}}$, we extract a set of corresponding pixel matches, denoted as  $\textbf{c}^{\text{ref}}, \ \textbf{c}^{\text{src}} \in \mathbb{R}^{N \times 2}$, using a dense matching model $M_{\theta}$ \cite{edstedt2023dkm},
\begin{equation}
\begin{aligned}\label{equation:prediction}
    \widehat{\textbf{P}}^\text{ref} = F_{\theta} (\textbf{I}^{\text{ref}})&,
    \ \
    \widehat{\textbf{P}}^{\text{src}} = F_{\theta} (\textbf{I}^{\text{src}}),
    \quad \\
    \boldsymbol{c}^{\text{ref}},\ \boldsymbol{c}^{\text{src}} =& M_{\theta}(\textbf{I}^{\text{ref}},\ \textbf{I}^{\text{src}}).
\end{aligned}
\end{equation}
In our scenario, we aim to register point clouds from each view while preserving consistency with the associated camera poses, which are either externally provided or estimated between views. 
Specifically, we consider two alternative cases: (1) aligning point maps using available input poses; and (2) aligning point maps after estimating the relative pose between views.
In both cases, the resulting pose priors are used to transform all point maps into a common coordinate system.
Unless otherwise noted, the point maps $\widehat{\textbf{P}}^{\text{ref}}$ and $\widehat{\textbf{P}}^{\text{src}}$ are assumed to be expressed in the common world coordinate after applying the corresponding rotation and translation.

%
\paragraph{Case 1: Alignment with Provided Camera Poses}
When the camera poses are available, the estimated point maps can be transformed into a common 3D coordinate, yielding $\widehat{\textbf{P}}$.
However, due to the inherent scale ambiguity of monocular cameras, the resulting point clouds are only accurate up to an unknown scale.
To address this ambiguity, MoGe \cite{wang2024moge} proposed a parallelized alignment solver that resolves affine transformation during the training of monocular point map estimation networks. 
Inspired by their approach, we extend the solver to operate across different views using $N$ pairs of corresponding points,
\begin{equation}\label{equation:solver1}
    (\alpha^{*}, \boldsymbol{\beta}^{*}) = \underset{\alpha, \boldsymbol{\beta}}{\text{argmin}} \sum_{(i, j) \in \mathcal{C}} \frac{1}{z_{i}} \| \alpha \widehat{\textbf{P}}^{\text{src}}_{j} + \boldsymbol{\beta} - \widehat{\textbf{P}}^{\text{ref}}_{i} \|_{1},
\end{equation}
where $\mathcal{C} = \{ ( \boldsymbol{c}^{\text{ref}}_{(n)},\ \boldsymbol{c}^{\text{src}}_{(n)} ) \}_{n=0}^{N-1}$ denotes the set of the matched points, $\alpha \in \mathbb{R}$ and $\boldsymbol{\beta} \in \mathbb{R}^3$ are the scale and shift parameters that produce the initially aligned point maps $\textbf{P}^{\text{ref}} = \widehat{\textbf{P}}^{\text{ref}}$ and $\textbf{P}^{\text{src}} = \alpha \widehat{\textbf{P}}^{\text{src}} + \boldsymbol{\beta}$.
$z_i$ represents the depth of the $i$-th reference point.

\paragraph{Case 2: Alignment without Provided Camera Poses}
MadPose \cite{yu2025relative} introduces a set of solvers that explicitly model affine corrections, using scale and shift parameters $\alpha, \beta^{\text{ref}}, \beta^{\text{src}} \in \mathbb{R}$ applied to monocular depth priors.
The solvers leverage matching points $(\boldsymbol{c}^{\text{ref}}, \boldsymbol{c}^{\text{src}})$ for relative pose estimation.
We use the solver under calibrated settings, where the intrinsic matrix $\boldsymbol{K}$ and the affine-invariant depth $\widehat{D}$ are derived from the predicted point maps $\widehat{\textbf{P}}$ by MoGe \cite{wang2024moge}.
MadPose utilizes a least square optimizer to compute scale and shift by triangulating and projecting corresponding depth points.
In this step, we observed that an excessively large predicted shift value can sometimes overwhelm the relative scale parameters, leading to an inaccurate depth map.
To improve stability, we compute the interquartile range (IQR) of valid depth values, defined as the spread between the 25th and 75th percentiles, and set an upper bound of $0.5 \times \text{IQR}$ on the shift term in Ceres solver \cite{agarwal2012ceres}.
\begin{equation}
\begin{aligned}\label{equation:solver2}
    \textbf{P}^{\text{ref}} &= \boldsymbol{K}^{-1}(\widehat{D}^{\text{ref}} + \beta^{\text{ref}})\widetilde{\boldsymbol{p}}^{\text{ref}},
    \\
    \textbf{P}^{\text{src}} &= \boldsymbol{K}^{-1}\left( \alpha (\widehat{D}^{\text{src}} + \beta^{\text{src}}) \right) \widetilde{\boldsymbol{p}}^{\text{src}},
\end{aligned}
\end{equation}
where $\boldsymbol{\widetilde{p}}$ denotes the homogeneous coordinates of the point map pixel $\boldsymbol{p}$.

\begin{figure*}[t]
    \centering
    \includegraphics[width=0.85\linewidth]{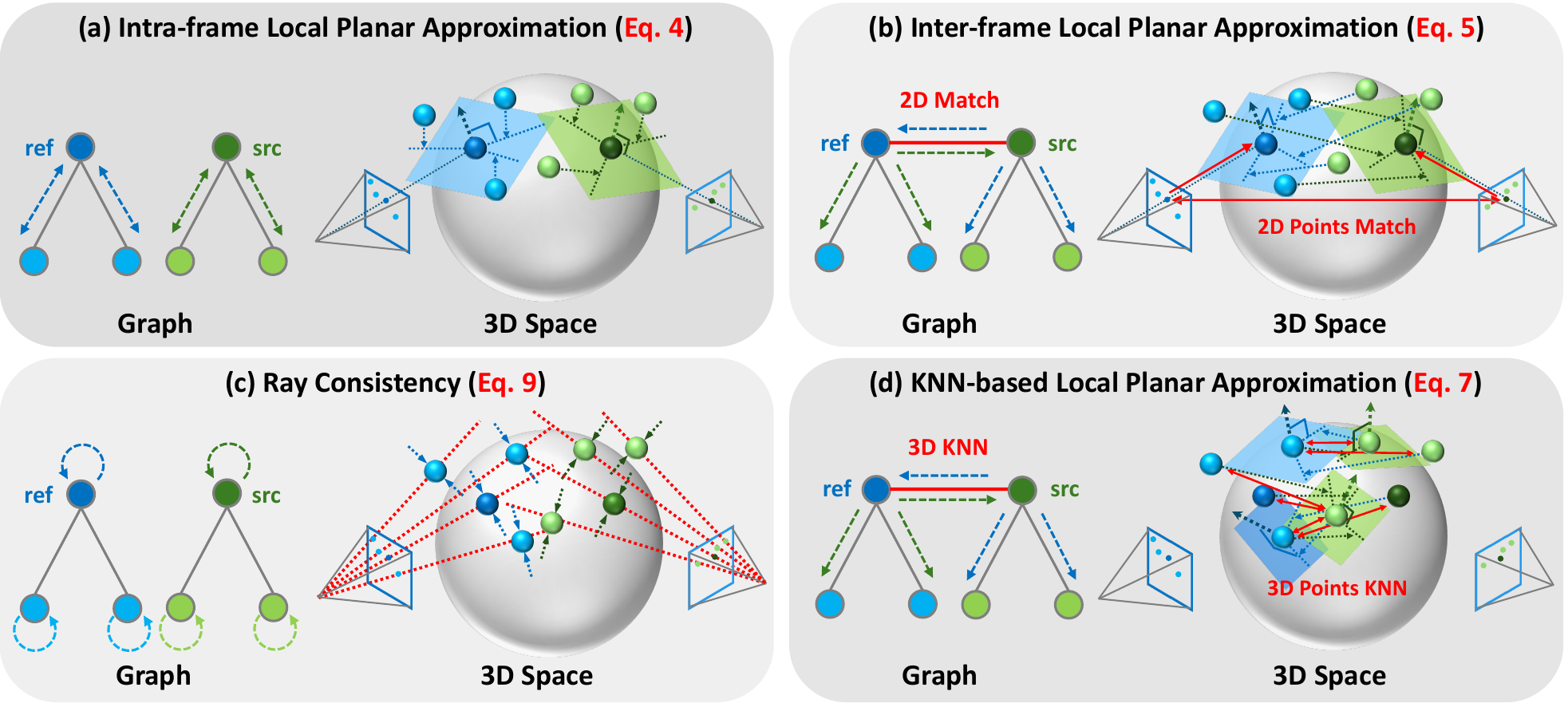}
    \vspace*{-2mm}
    \caption{
    {\bf Illustration of the proposed geometric constraints for the graph optimization.}
    \textbf{Graph} depicts the abstract graph structure, where nodes indicate 3D points and edges with dotted lines indicate geometric constraints.
    \textbf{3D Space} shows the corresponding spatial relationships of the graph structure, where colored rectangles represent local tangent planes and small spheres denote 3D points.
    Each subfigure presents a distinct type of geometric constraint incorporated into our optimization:
    \textcolor{OliveGreen}{\textbf{(a)}} Enforces local surface smoothness within the same frame by assuming neighboring 3D points lie on a shared local plane. This regularization is applied within each individual view.
    \textcolor{OliveGreen}{\textbf{(b)}} Propagates geometric smoothness across frames using 2D point correspondences. Matched points across views are encouraged to lie on a consistent local plane, supporting cross-view surface coherence.
    \textcolor{OliveGreen}{\textbf{(c)}} Ensures that 3D points align with the viewing rays of their corresponding pixels.
    Ray consistency allows reprojection consistency of the 3D points within each frame.
    \textcolor{OliveGreen}{\textbf{(d)}} Applies local surface smoothness constraints across views using 3D K-nearest neighbors (KNN).
    This provides additional regularization for corresponding points that were not detected by the 2D matcher, helping to achieve more complete alignment. 
    %
    %
    }
    \label{fig:graph}
    \vspace*{-4mm}
\end{figure*}

\subsection{Geometric Constraints and Refinement}
\label{section:refinement}
As shown in Fig. \ref{fig:toy} and \ref{fig:pipeline}, the initial affine alignment yields a coarse registration of point maps across different views.
However, residual discrepancies still remain between the matched 3D points.
Here, we propose a graph-based, locally planar approximation to refine the point maps with precise geometric awareness and view consistency.
A plane $\boldsymbol{\mathcal{P}}$ can be defined by a certain point $\textbf{P}_0$ and its corresponding normal vector $\boldsymbol{n}_0$.
The normal map $\boldsymbol{n}$ is computed by performing cross products between neighboring points in the point map.
For any point $\textbf{P}_{x}$ in the point map lying on plane $\boldsymbol{\mathcal{P}}$, the following condition holds: $\boldsymbol{n}_0 \cdot (\textbf{P}_x - \textbf{P}_0) = 0$.
Based on this simple equation, Rossi et al. \cite{rossi2020joint} proposed a monocular depth refinement method using piecewise optimization, focusing on improving geometric awareness within a single view. 
In contrast, we address 3D point consistency across multiple viewpoints, explicitly enforcing geometric consistency between views.

Figure \ref{fig:graph} visualizes the geometric constraints.
We define the 3D points as nodes and assign edge weights $w$ based on the likelihood that two neighboring points lie on the same plane. 
For simplicity, we formulate the constraints with respect to the reference frame; the same formulation can be applied to the source frame in a symmetric manner.
For each point map, we define an intra- and inter-frame loss function based on the local planar approximation,

{\footnotesize
\begin{equation}
\begin{aligned}\label{equation:plane_intra}
    \!L_{intra}^{\text{ref}} \!\!= \!\!\sum_{i \in \vphantom{j)} \ \Omega\vphantom{12}} \sum_{i' \sim i} w^{\text{2D}}_{ii'} \!\!\left(\! \| \boldsymbol{n}^{\text{ref}}_i \!\cdot\! (\textbf{P}^{\text{ref}}_{i'} \!-\! \textbf{P}^{\text{ref}}_{i}) \|_2 \ \!+\! \gamma \ \| \boldsymbol{n}^{\text{ref}}_{i'} \!-\! \boldsymbol{n}^{\text{ref}}_{i} \|_2 \!\right),
\end{aligned}
\end{equation}
}
{\footnotesize
\begin{equation}
\begin{aligned}
    \!L_{inter}^{\text{ref}} \!\!=\!\!\!\!\! \sum_{(i, j) \in \mathcal{C} } \!\sum_{j' \sim j} &w^{\text{2D}}_{jj'} \left(\! \| \boldsymbol{n}^{\text{ref}}_{i} \!\cdot\! (\textbf{P}^{\text{src}}_{j'} \!-\! \textbf{P}^{\text{ref}}_{i}) \|_2  \!+\! \gamma\! \| \boldsymbol{n}^{\text{src}}_{j'} \!-\! \boldsymbol{n}^{\text{ref}}_{i} \|_2 \!\right),
    \\
    + \rho\!\!\! \sum_{\!\!\!(i, j) \in \mathcal{C}} \!\sum_{\substack{(i'\!,j'\!) \!\sim\! (i, j)}} &\!\!\!\!\!\!\!\!w^{\text{2D}}_{ii'}w^{\text{2D}}_{jj'} \!\!\left(\!\| \boldsymbol{n}^{\text{ref}}_{i} \!\cdot\! (\textbf{P}^{\text{ref}}_{i'} \!-\! \textbf{P}^{\text{src}}_{j'}) \!\|_2 \!+\! \frac{\gamma}{2}\! \| \boldsymbol{n}^{\text{ref}}_{i} \!-\! \boldsymbol{n}^{\text{src}}_{j} \!\|_2\!\right)
    \label{equation:plane_inter}
\end{aligned}
\end{equation}
}
\!\!where $i' \!\sim \!i$ and $j' \!\sim\! j$ denote the neighboring pixels of $i$ and $j$ in the graph, and $w^{\text{2D}}_{ii'}$ and $w^{\text{2D}}_{jj'}$ represent the edge weights between $i$ and $i'$, and between $j$ and $j'$.
$\Omega$ indicates the valid pixel regions of the frame. 
We set $\gamma$ as 0.5 and $\rho$ as 0.1.
Using the corresponding points $\mathcal{C}$ from 2D matching \cite{edstedt2023dkm}, we define the geometric relationship between the different view frames.
If available, we apply the RANSAC algorithm to select inliers in pairs of matches $\mathcal{C}$.
In Eq. \ref{equation:plane_intra} and \ref{equation:plane_inter}, the first term inside the parentheses enforces coplanarity between two points, while the second term promotes planar consistency among neighboring points.
Assuming that areas with similar textures lie on the same surface \cite{krahenbuhl2011efficient, foi2012foveated, rossi2018nonsmooth, rossi2020joint}, edge weights $w^{\text{2D}}_{ll'}$ are defined using local patch similarity and the spatial distance between pixels,
%
{
\small
\begin{equation}\label{equation:w_2d_general}
\begin{aligned}
    w^{\text{2D}}_{ll'} =
    &\exp\left(
        -\frac{
            \| \mathbf{Q}^{f}_l - \mathbf{Q}^{f}_{l'} \|^2_F
        }{
            2\sigma^2_{\text{int}}
        }
    \right)
    \exp\left(
        -\frac{
            \| l - l' \|^2_2
        }{
            2\sigma^2_{\text{spa}}
        }
    \right),
    \\
    &\quad \quad \text{where} \ \ l=
    \begin{cases}
        i \quad \text{with} \ \  f=\text{ref} \\
        j \quad \text{with} \ \  f=\text{src}
    \end{cases} \ ,
\end{aligned}
\end{equation}
}
\!\!\!where $\textbf{Q}^{f}_{l}$ represents a patch centered at pixel \!$l$ in image frame $\textbf{I}^{f}$\!, $\| \cdot \|_{F}$ denotes the Frobenius norm, and $\sigma_{\text{int}}$ and $\sigma_{\text{spa}}$ are set to 0.07 and 3.0, respectively. 
%
%
%
%
Although the inter-frame relationship is defined in Eq. \ref{equation:plane_inter}, the corresponding points $\mathcal{C}$ and their neighboring pixels cover only part of the images, causing performance to vary depending on the number of matches.
Therefore, we perform an additional k-nearest neighbor (kNN) search on 3D points, using $\textbf{P}^{\text{ref}}$ as the query set and $\textbf{P}^{\text{src}}$ as the support set, retrieving the indices $j$ in the set of nearest neighbors $\mathcal{N}_k (i)$ that minimize $\| \textbf{P}^{\text{ref}}_{i} - \textbf{P}^{\text{src}}_{j} \|_2$,
\begin{equation}\label{plane_3D}
\begin{aligned}
    L_{knn}^{\text{ref}} = \sum_{i \in \Omega} \sum_{j \in \mathcal{N}_k (i)} w^{\text{3D}}_{ij} \Big( \| \boldsymbol{n}^{\text{ref}}_{i} \cdot (\textbf{P}^{\text{ref}}_{i} - \textbf{P}^{\text{src}}_{j}) \|_2 \ \\
    + \ \| \boldsymbol{n}^{\text{src}}_{j} \cdot (\textbf{P}^{\text{src}}_{j} - \textbf{P}^{\text{ref}}_{i}) \|_2 \ + \ \| \boldsymbol{n}^{\text{ref}}_{i} - \boldsymbol{n}^{\text{src}}_{j} \|_2 \Big).
\end{aligned}
\end{equation}
%
The edge weight $w^{\text{3D}}_{ij}$ is designed to add connections between points from different views that lack feature matches in the images. 
Since distance information is already embedded in the kNN process, we employ normal similarity instead of spatial distance in this term,
\begin{equation}
    w^{\text{3D}}_{ij} = \text{exp} \left( -\frac{\|\textbf{I}^{\text{ref}}_{i} - \textbf{I}^{\text{src}}_{j} \|^2_2}{2 \sigma_{\text{int}}^2} \right) \text{exp}\left( -\frac{\|\boldsymbol{n}^{\text{ref}}_i - \boldsymbol{n}^{\text{src}}_j \|^2_2}{2 \sigma_{\text{int}}^2} \right).
\end{equation}
Throughout this refinement, we parameterize the point map instead of the depth map to enable more flexible optimization.
When the depth map is used as the parameter, the point cloud can only move along the viewing ray direction, limiting the optimization.
In contrast, by directly optimizing the point cloud, we allow adjustments in all $xyz$ directions.
To prevent the optimized points from drifting too far from their original pixel positions, we add a constraint that minimizes the distance between each point $\textbf{P}$ and its corresponding viewing ray $\boldsymbol{r}$ during the graph optimization process,
\begin{equation}\label{equation:ray}
    L_{r}^{\text{ref}} = \sum_{i \in \Omega} \| \boldsymbol{r}^{\text{ref}}_{i} \times \textbf{P}^{\text{ref}}_{i} \|_2.
\end{equation}
To avoid trivial solutions during graph optimization, we impose two regularization terms based on the original point and normal data. 
We leverage the original structural information as a prior to encourage the refined point maps to maintain a similarity transformation (under fixed poses) with the original points $\overline{\textbf{P}}$, incorporating an additional scale parameter $s \in \mathbb{R}$. 
We also enforce that the refined normals do not deviate significantly from the input normal maps $\overline{\boldsymbol{n}}$,

{
\footnotesize
\begin{equation}\label{equation:rigid}
    \!\!\!L_s^{\text{ref}} = \sum_{i \in \Omega} \left\| \| \textbf{P}^{\text{ref}}_{i} \!-\! \frac{1}{|\Omega|} \!\!\sum_{k \in \Omega} \overline{\textbf{P}}^{\text{ref}}_{k} \|_2 \!-\! s\| \overline{\textbf{P}}^{\text{ref}}_{i} \!\!-\! \frac{1}{| \Omega |} \sum_{k \in \Omega} \!\overline{\textbf{P}}^{\text{ref}}_{k} \|_2 \right\|_1 \!\!\!\textbf{m}^{\text{ref}}_{i}\!,
\end{equation}
}
{
\footnotesize
\begin{equation}\label{equation:normal}
    L_n^{\text{ref}} = \sum_{i \in \Omega} \| \boldsymbol{n}^{\text{ref}}_{i} - \boldsymbol{\overline{n}}^{\text{ref}}_{i} \|_2 \ \textbf{m}^{\text{ref}}_i.
\end{equation}
}
\!\!where the confidence mask $\textbf{m}^{\text{ref}}$ is obtained from \cite{wang2024moge}. 
The total loss function is formulated as a weighted sum of the proposed loss terms,
\begin{equation}\label{equation:total}
\begin{aligned}
    L_{\text{total}} = &\sum_{v \in \{\text{ref},\ \text{src}\}} \lambda_p \left( L^{v}_{intra} \ + \ L^{v}_{inter} \ + \ L^{v}_{knn} \right) \\
    &\quad+ \lambda_r L^{v}_{r} \ \ + \ \ \lambda_s L^{v}_{s} \ \ + \ \  \lambda_n L^{v}_{n},
\end{aligned}
\end{equation}
where $\lambda_p$, $\lambda_r$, $\lambda_s$, and $\lambda_n$ are set to 30, 50, 0.1, and 10, respectively.
\begin{table*}
\begin{center}
\renewcommand\arraystretch{1.2}
\setlength{\tabcolsep}{1pt} 
\small
\hspace{-3mm}
\resizebox{\textwidth}{!}{
\begin{tabular}{llccccrrrrrrrrrrrr}
 \hline
\specialrule{1.5pt}{0.5pt}{0.5pt}
\multicolumn{2}{l}{\multirow{2}{*}{Methods}} & GT & GT & GT & Scaling & \multicolumn{2}{c}{KITTI} & \multicolumn{2}{c}{ScanNet} & \multicolumn{2}{c}{ETH3D} & \multicolumn{2}{c}{DTU} & \multicolumn{2}{c}{T\&T} & \multicolumn{2}{c}{Average} \\
\cline{7-18}
 && Pose & Range & Intrinsics &  & rel $\downarrow$ & $\tau \uparrow$ & rel $\downarrow$ & $\tau \uparrow$ & rel $\downarrow$ & $\tau \uparrow$ & rel $\downarrow$ & $\tau \uparrow$ & rel$\downarrow$ & $\tau \uparrow$ & rel$\downarrow$ & $\tau \uparrow$ \\
\specialrule{1.5pt}{0.5pt}{0.5pt}
\multirow{2}{*}{(a)} & COLMAP~\cite{schonberger2016structure, schonberger2016pixelwise} & $\checkmark$ & $\times$ & $\checkmark$ & $\times$ &{\bf 12.0}&{\bf 58.2} &{\bf 14.6}&{\bf 34.2}&{\bf 16.4}&{\bf 55.1}&{\bf 0.7}&{\bf 96.5}&{\bf 2.7}& 95.0 &{\bf 9.3} & {\bf 67.8} \\
&COLMAP Dense~\cite{schonberger2016structure, schonberger2016pixelwise} & $\checkmark$&$\times$ & $\checkmark$ & $\times$ & 26.9 & 52.7 & 38.0 & 22.5 & 89.8 & 23.2 & 20.8 & 69.3 & 25.7 & 76.4 & 40.2 & 48.8 \\
\hline
\multirow{5}{*}{(b)} & MVSNet~\cite{yao2018mvsnet} & $\checkmark$ & $\checkmark$ &$\checkmark$ & $\times$ & 22.7 & 36.1 & 24.6 & 20.4 & 35.4 & 31.4 & (1.8) & $(86.0)$ & 8.3 & 73.0 & 18.6 & 49.4 \\
& MVSNet Inv. Depth~\cite{yao2018mvsnet} & $\checkmark$ & $\checkmark$ &$\checkmark$ & $\times$ & 18.6 & 30.7 & 22.7 & 20.9 & 21.6 & 35.6 & (1.8) & $(86.7)$ & 6.5 & 74.6 & 14.2 & 49.7 \\
& Vis-MVSNet~\cite{zhang2023vis} & $\checkmark$ & $\checkmark$ & $\checkmark$ & $\times$ &{\bf 9.5}&{\bf 55.4}& 8.9 & 33.5 &{\bf 10.8}&{\bf 43.3}&{\bf (1.8)} &{\bf (87.4)} &{\bf 4.1}&{\bf 87.2} &{\bf 7.0} &{\bf 61.4} \\
& MVS2D ScanNet~\cite{yang2022mvs2d} & $\checkmark$ & $\checkmark$ & $\checkmark$ & $\times$ & 21.2 & 8.7 & (27.2) & (5.3) & 27.4 & 4.8 & 17.2 & 9.8 & 29.2 & 4.4 & 24.4 & 6.6 \\
& MVS2D DTU~\cite{yang2022mvs2d} & $\checkmark$ & $\checkmark$& $\checkmark$ & $\times$ & 226.6 & 0.7 & 32.3 & 11.1 & 99.0 & 11.6 & (3.6) & (64.2) & 25.8 & 28.0 & 77.5 & 23.1 \\
\hline
\multirow{10}{*}{(c)} & DeMon~\cite{ummenhofer2017demon} & $\checkmark$ & $\times$ &$\checkmark$ & $\times$ & 16.7 & 13.4 & 75.0 & 0.0 & 19.0 & 16.2 & 23.7 & 11.5 & 17.6 & 18.3 & 30.4 & 11.9 \\
& DeepV2D KITTI~\cite{teed2018deepv2d} & $\checkmark$ & $\times$ &$\checkmark$ & $\times$ & (20.4) & (16.3) & 25.8 & 8.1 & 30.1 & 9.4 & 24.6 & 8.2 & 38.5 & 9.6 & 27.9 & 10.3 \\
& DeepV2D ScanNet~\cite{teed2018deepv2d} & $\checkmark$ & $\times$ &$\checkmark$ & $\times$ & 61.9 & 5.2 & (3.8) & (60.2) & 18.7 & 28.7 & 9.2 & 27.4 & 33.5 & 38.0 & 25.4 & 31.9 \\
& MVSNet~\cite{yao2018mvsnet} & $\checkmark$ & $\times$ &$\checkmark$ & $\times$ & 14.0 & 35.8 & 1568.0 & 5.7 & 507.7 & 8.3 & (4429.1) & (0.1) & 118.2 & 50.7 & 1327.4 & 20.1 \\
& MVSNet Inv. Depth~\cite{yao2018mvsnet} & $\checkmark$ & $\times$ &$\checkmark$ & $\times$ & 29.6 & 8.1 & 65.2 & 28.5 & 60.3 & 5.8 & (28.7) & (48.9) & 51.4 & 14.6 & 47.0 & 21.2 \\
& Vis-MVSNet \cite{zhang2023vis} & $\checkmark$ & $\times$ &$\checkmark$ & $\times$ & 10.3 &{\bf 54.4}& 84.9 & 15.6 & 51.5 & 17.4 & (374.2) & (1.7) & 21.1 & 65.6 & 108.4 & 31.0 \\
& MVS2D ScanNet~\cite{yang2022mvs2d} & $\checkmark$ &$\times$ &$\checkmark$ &$\times$ & 73.4 & 0.0 & (4.5) & (54.1) & 30.7 & 14.4 & 5.0 & 57.9 & 56.4 & 11.1 & 34.0 & 27.5 \\
& MVS2D DTU~\cite{yang2022mvs2d} &$\checkmark$ & $\times$ & $\checkmark$ & $\times$ & 93.3 & 0.0 & 51.5 & 1.6 & 78.0 & 0.0 & (1.6) & (92.3) & 87.5 & 0.0 & 62.4 & 18.8 \\
& Robust MVD Baseline~\cite{schroppel2022benchmark} &$\checkmark$&$\times$&$\checkmark$&$\times$ & 7.1 & 41.9 & 7.4 & 38.4& 9.0& 42.6 &{\bf 2.7}&{\bf 82.0}& 5.0&{\bf 75.1}& 6.3 & 56.0 \\
& {\bf MoRe (ours)} &$\checkmark$&$\times$&$\checkmark$&med& {\bf 6.17}& 37.96& {\bf 3.47}& {\bf 66.51}& {\bf 4.20}& {\bf 64.46}& 3.67& 75.47& {\bf 3.21}& 71.47& {\bf 4.04}& {\bf 63.82} \\
\hline
\multirow{5}{*}{(d)} & DeMoN~\cite{ummenhofer2017demon} &$\times$&$\times$ &$\checkmark$& $\|\mathbf{t}\|$ & 15.5 & 15.2 & 12.0 & 21.0 & 17.4 & 15.4 & 21.8 & 16.6 & 13.0 & 23.2 & 16.0 & 18.3 \\
& DeepV2D KITTI~\cite{teed2018deepv2d} &$\times$&$\times$&$\checkmark$&med& (3.1) & (74.9) & 23.7 & 11.1 & 27.1 & 10.1 & 24.8 & 8.1 & 34.1 & 9.1 & 22.6 & 22.7 \\
& DeepV2D ScanNet~\cite{teed2018deepv2d} &$\times$&$\times$&$\checkmark$& med &10.0 & 36.2 &(4.4) & (54.8) & 11.8 & 29.3 & 7.7 & 33.0 & 8.9 & 46.4 & 8.6 & 39.9 \\
& DUSt3R \cite{wang2024dust3r} &$\times$&$\times$&$\times$&med&9.11& 39.49& (4.93)&(60.20) &{\bf 2.91}&{\bf 76.91}& 3.52&69.33 & 3.17&{\bf 76.68}&4.73&{\bf 64.52} \\
& {\bf MoRe (ours)} &$\times$&$\times$&$\times$&med& {\bf 5.40}& {\bf 43.05}& {\bf 3.49}& {\bf 66.15}& 3.82& 68.12& {\bf 3.12}& {\bf 70.29}& {\bf 3.03}& 73.78& {\bf 3.74}& 64.48 \\
%
\specialrule{1.5pt}{0.5pt}{0.5pt}
\end{tabular}}
\vspace*{-2mm}
\normalsize
\caption{
\textbf{Multi-view Depth Evaluation} under different settings on the benchmark dataset \cite{schroppel2022benchmark}: 
(a) Classical methods using ground truth poses and intrinsics;
(b) Learning-based methods with ground truth poses, intrinsics, and depth ranges;
(c) Learning-based methods with poses and intrinsics but without depth ranges. Only MoRe (ours) is optimization-based.
(d) Methods without access to ground truth poses or depth ranges.
Methods marked with (parentheses) are trained on data from the same domain. The best results are shown in \textbf{bold}.
}
\label{tab:mvd}
\end{center}
\vspace*{-4mm}
\end{table*}

\section{Experiments}
\label{sec:experiments}

\subsection{Experiments Settings}
\paragraph{Dataset}
According to \cite{schroppel2022benchmark}, we compare our method with other multi-view depth estimation methods.
We utilize the DTU \cite{aanaes2016large}, ETH3D \cite{schops2017multi}, Tanks and Temples \cite{knapitsch2017tanks}, ScanNet \cite{dai2017scannet}, and KITTI \cite{geiger2013vision} datasets to evaluate geometry estimation performance.
All test images are uniformly resized such that the longer side is scaled to 512 pixels while maintaining their original aspect ratio.
Additionally, to evaluate novel-view rendering in sparse-view scenarios, we use seven scenes from the Tanks and Temples dataset \cite{knapitsch2017tanks}, with view counts ranging from 3 to 12, following the protocol of InstantSplat \cite{fan2024instantsplat}.

\paragraph{Implementation Details}
We employ the Adam optimizer \cite{kingma2014adam} for gradient-based optimization and accelerate convergence using a multi-scale strategy, similar to the approach in \cite{rossi2020joint}. 
At each scale level $l \in \{0, \dots, L-1\}$, the point map $\textbf{P} \in \mathbb{R}^{\lfloor W / 2^l \rfloor \times \lfloor H / 2^l \rfloor \times 3}$ is progressively downsampled by a factor of $2^l$. 
This downsampling not only reduces the number of nodes to accelerate optimization, but also encourages the model to capture relationships between more distant nodes, effectively expanding the receptive field.
For our experiments, we set $L = 2$ and apply learning rates of $5 \times 10^{-3}$ at each level. 
The optimization is carried out for 50 and 50 iterations at levels 0, 1, respectively.
For a fair comparison, Section \ref{subsec:3drecon} and \ref{subsec:nvs}, where ground-truth poses are available, adopt the initial alignment method described in Case 1 of Sec. \ref{section:initial_align}.
Section \ref{subsec:mvd} reports the performance of both Case 1 and Case 2.

\subsection{Multi-view Depth}
\label{subsec:mvd}
\begin{figure*}[t!]
    \centering
    \includegraphics[width=0.95\linewidth]{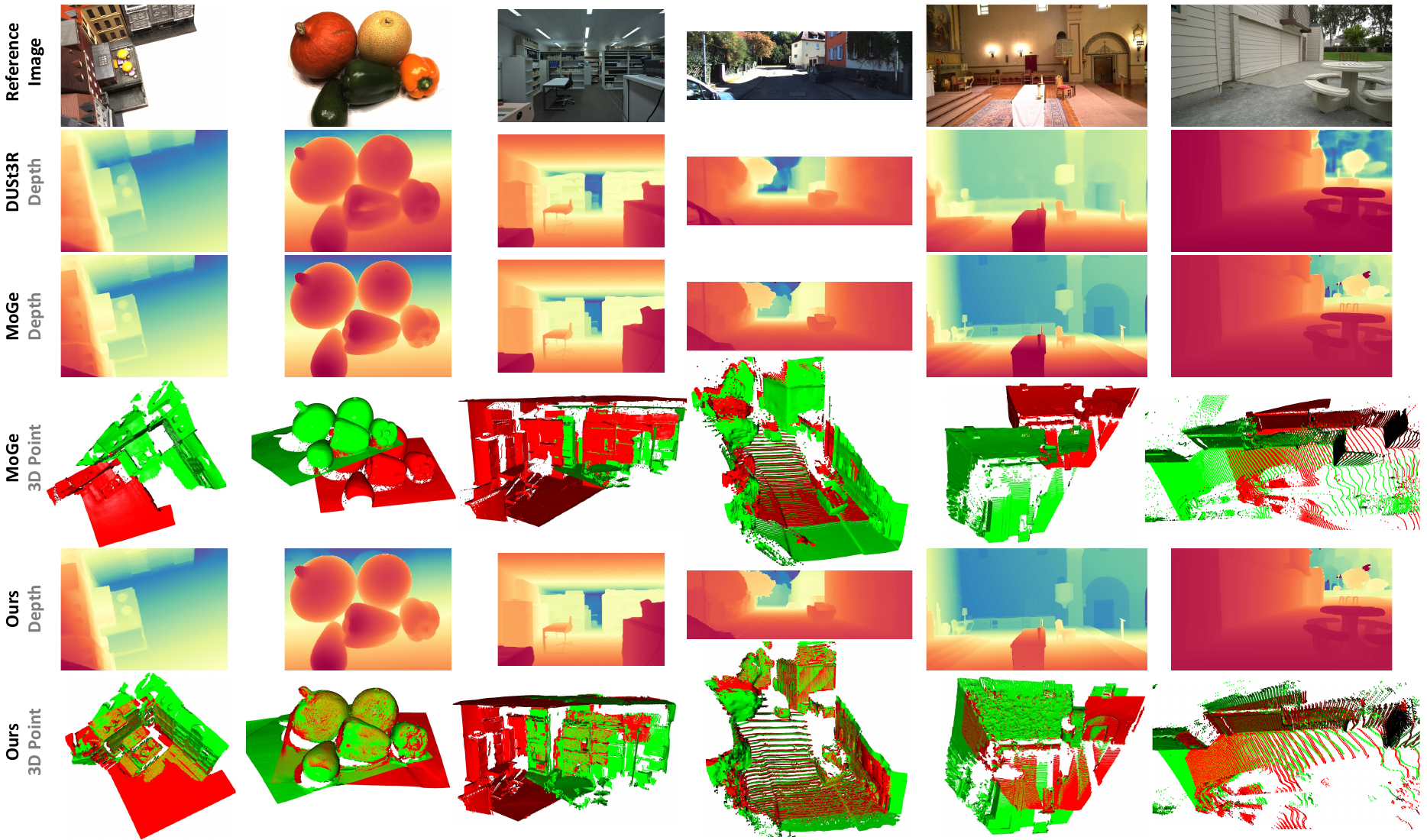}
    \vspace*{-2mm}
    \caption{
    \textbf{Qualitative Results on Depth Estimation and 3D Reconstruction.}
    The first row shows the reference images.
    DUSt3R \cite{wang2024dust3r}, a multi-view 3D foundation model, estimates depth using additional source frames, while MoGe \cite{wang2024moge} performs monocular depth estimation independently for each frame. 
    As a result, MoGe produces detailed depth maps but generates misaligned point clouds across views due to inconsistent scale. 
    In contrast, our method achieves similarly detailed depth predictions while producing 3D reconstructions with consistent scale and alignment across frames.
    Red and green points indicate point clouds generated from the reference and source views, respectively.
    }
    \label{fig:mvd}
\end{figure*}

We evaluate our method on the task of multi-view stereo depth estimation. 
After performing affine refinement to align multiple monocular pointmaps, depth values are predicted by simply selecting the z-coordinates of the estimated 3D points. 
Following the evaluation protocol of \cite{schroppel2022benchmark}, we assess performance on five standard benchmarks: KITTI \cite{geiger2013vision}, DTU \cite{aanaes2016large}, ETH3D \cite{schops2017multi}, Tanks and Temples \cite{knapitsch2017tanks}, and ScanNet \cite{dai2017scannet}. 
We report Absolute Relative Error (Rel) and Inlier Ratio ($\gamma$) with a threshold of 1.03 for each test set, as well as the average performance across all datasets.
Figure \ref{fig:mvd} illustrates that MoGe \cite{wang2024moge}, trained on a broader set of monocular data, produces more detailed depth estimates compared to the stereo-based method DUSt3R \cite{wang2024dust3r}.
With our proposed MoRe method, which applies affine transformation and graph optimization to the pointmaps predicted by MoGe, the original structure is preserved.
As a result, although the predicted depth maps from MoGe and MoRe appear visually similar, the refined 3D points produced by MoRe exhibit improved consistency across views.
As shown in Table \ref{tab:mvd}, our method achieves superior or comparable performance to previous methods.
During evaluation, when the estimated points are only defined up to scale, we apply median scaling to enable quantitative comparisons.

\subsection{3D Reconstruction}
\label{subsec:3drecon}
To evaluate 3D reconstruction performance, we conduct experiments on the DTU dataset \cite{aanaes2016large}.
Since our method relies on monocular geometry estimation as an initial prior, which inherently suffers from scale ambiguity, additional post-processing is required to align the estimated 3D structure with the ground truth coordinate system.
We select a central frame as the reference view and treat the remaining frames as source views.
During the MoRe process, we estimate an affine transformation to align the point maps and camera positions between the reference frame and the source frames.
While the resulting pointmaps are consistent across views, they are not yet aligned to the ground truth coordinate system required for metric evaluation.
To address this, we align the estimated camera positions with the ground truth camera origins using Eq. \ref{equation:solver1}.
Table \ref{tab:mvs_dtu} shows the averaged accuracy, averaged completeness, and overall average error, following the evaluation protocol of DUSt3R \cite{wang2024dust3r}.
Similar to DUSt3R, our method does not rely on subpixel accurate triangulation or training specifically on the DTU dataset, and is evaluated in a zero shot setting. 
As a result, it does not achieve the best performance.
Nevertheless, the results show that our monocular approach performs comparably to DUSt3R, which operates in a multiview setting, and even outperforms it when ground truth camera poses are available.

\begin{table}[t]
    \centering
    \resizebox{0.85\linewidth}{!}{
    \begin{tabular}{llcccc}
    \toprule
    & Methods & GT cams & Acc.$\downarrow$ & Comp.$\downarrow$ & Overall$\downarrow$       \\
    \hline
    \multirow{4}{*}{(a)} & Camp~\cite{campbell2008using} & $\checkmark$ & 0.835 & 0.554 & 0.695 \\
    &Furu~\cite{furukawa2009accurate} & $\checkmark$ & 0.613 & 0.941 & 0.777 \\
    &Tola~\cite{tola2012efficient} & $\checkmark$ & 0.342 & 1.190 & 0.766 \\
    &Gipuma~\cite{galliani2015massively} & $\checkmark$ &\textbf{ 0.283} & 0.873 & 0.578\\
    \hline
    \multirow{9}{*}{(b)} &MVSNet~\cite{yao2018mvsnet} & $\checkmark$ &0.396 & 0.527 & 0.462 \\
    &CVP-MVSNet~\cite{Yang_2020_CVPR} & $\checkmark$ & 0.296 & 0.406 & 0.351 \\
    &UCS-Net~\cite{cheng2020deep} & $\checkmark$ & 0.338 & 0.349 & 0.344 \\
    & CER-MVS~\cite{ma2022multiview} & $\checkmark$ & 0.359 & 0.305 & 0.332 \\
    & CIDER~\cite{xu2020learning} & $\checkmark$ & 0.417 & 0.437 & 0.427 \\
    & CasMVSNet~\cite{gu2020cascade} & $\checkmark$ & 0.325 & 0.385 & 0.355 \\
    & PatchmatchNet~\cite{wang2021patchmatchnet} & $\checkmark$ & 0.427 & 0.277 & 0.352 \\
    & GeoMVSNet~\cite{zhang2023geomvsnet} & $\checkmark$ & 0.331 & \textbf{0.259} & \textbf{0.295} \\
    \hline
    \multirow{3}{*}{(c)}& DUSt3R \cite{wang2024dust3r} & $\times$ &   2.677  &  0.805  & 1.741  \\
    & DUSt3R \cite{wang2024dust3r} & $\checkmark$ & 3.654& 4.994& 4.324\\ 
    & {\bf MoRe (ours)} & $\checkmark$ & 2.202 &	1.352 &	1.777  \\
    \bottomrule
    \end{tabular}
    }
    \vspace*{2mm}
\normalsize
\caption{
\textbf{Multi-view Stereo} results on the DTU dataset \cite{aanaes2016large}.
(a) Traditional triangulation-based methods,
(b) Learning-based methods trained specifically on DTU,
and (c) Zero-shot evaluation results without specific training on DTU.
Our method (MoRe) achieves competitive performance and outperforms DUSt3R when ground-truth camera parameters are provided.
\label{tab:mvs_dtu}
}
\end{table}

\begin{figure}[t]
    \centering
    \includegraphics[width=1.\linewidth]{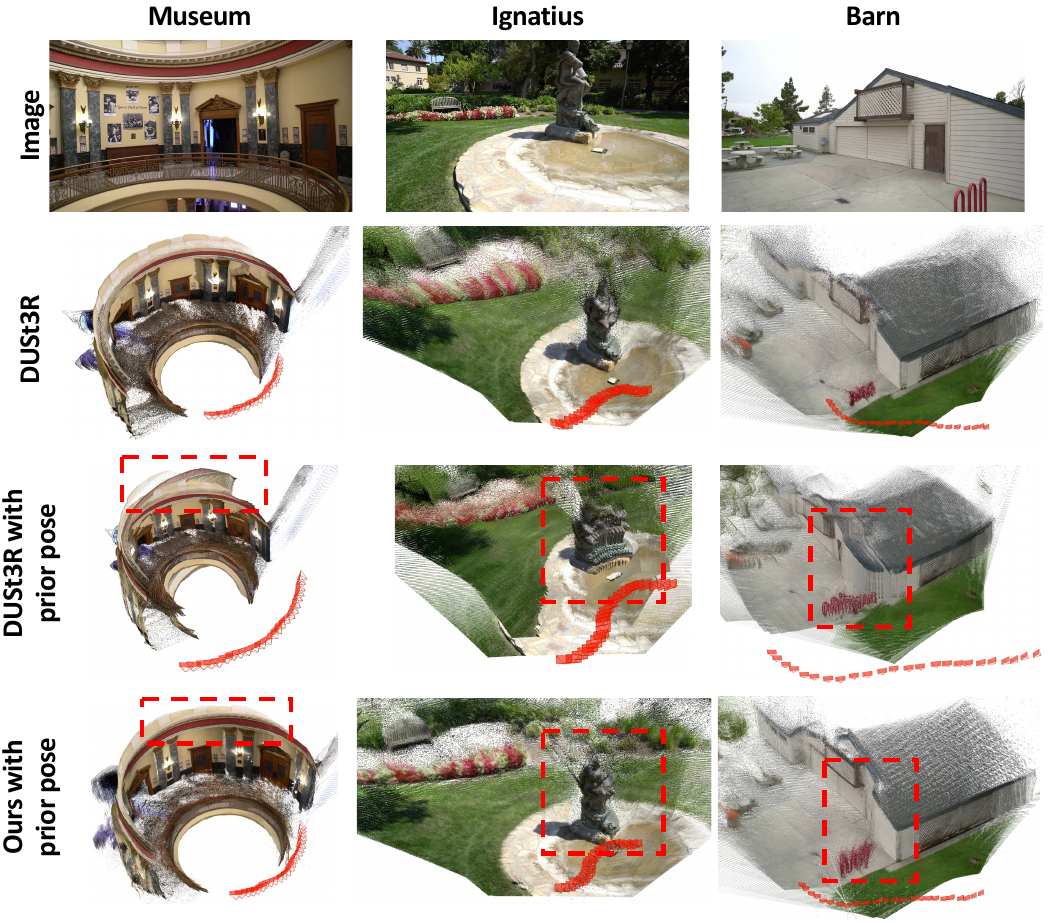}
    \vspace*{-6mm}
    \caption{
    \textbf{Qualitative Comparison of Global Alignment.}
    Given prior camera poses, DUSt3R \cite{wang2024dust3r} exhibits noticeable misalignments in 3D reconstructions (highlighted in red boxes), such as duplicated structures.
    In contrast, our method produces more globally consistent point clouds using the same camera poses. 
    This demonstrates that monocular geometry can be effectively aligned through our refinement method.
    }
    \label{fig:global_align}
    \vspace*{-4mm}
\end{figure}


\subsection{Ablation Study}
Table \ref{tab:ablation} demonstrates the effectiveness of our method for monocular geometry alignment.
Applying affine transformation between views improves the accuracy of monocular geometry by leveraging geometric information from other views.
Furthermore, incorporating graph-based optimization with local planar approximation using surface normals leads to more accurate geometry across views.

\begin{table}[t]
    \centering
    \resizebox{0.85\linewidth}{!}{
    \begin{tabular}{l|c|cc}
    \toprule
    Methods & Alignment & rel$\downarrow$ & $\tau \uparrow$ \\
    \hline
    MoGe \cite{wang2024moge} & $\times$ & 3.91 & 63.84 \\
    MoRe-align & Affine Transformation & 3.78 & 64.15 \\
    MoRe-full & Graph Optimization & \textbf{3.74} & \textbf{64.48} \\
    \bottomrule
    \end{tabular}
    }
    \vspace*{-2mm}
\normalsize
\caption{
\textbf{Ablation Study} for the proposed method on the benchmark dataset \cite{schroppel2022benchmark}.
MoRe-align indicates initial affine alignment from MoGe \cite{wang2024moge} predictions, while MoRe-full incorporates graph optimization for pixel-level cross-view point alignment.
Our full method achieves the best performance on both metrics.
\label{tab:ablation}
}
\end{table}

%
%
    
    



\subsection{Novel View Synthesis}
\label{subsec:nvs}
We evaluate the pointmaps refined by our method through downstream novel view synthesis tasks on the Tanks and Temples dataset.
The original 3D Gaussian Splatting (3DGS) \cite{kerbl20233d} relies on sparse points initialized from Structure-from-Motion (SfM) \cite{schonberger2016structure}, followed by a densification step that increases the number of Gaussians to cover both under- and over-reconstructed regions.
However, in sparse-view settings, this straightforward strategy struggles due to poor initializations from SfM, often resulting in overfitting to the training views.
Several prior works \cite{zhu2024fsgs,liu2024mvsgaussian,kotovenko2025edgs} have already highlighted the importance of dense initialization when applying 3DGS in sparse-view scenarios.
In particular, EDGS \cite{kotovenko2025edgs} bypasses incremental densification and instead uses dense feature matching to obtain a more reliable dense initialization.
Inspired by EDGS \cite{kotovenko2025edgs}, we similarly skip the densification step and directly optimize Gaussian splats for evaluating point maps for novel view synthesis task.
MoRe generates globally aligned point clouds for initialization, as shown in Fig. \ref{fig:global_align}.

We report PSNR, SSIM \cite{wang2004image}, and LPIPS \cite{zhang2018unreasonable} for the full images to evaluate rendering quality. 
DUSt3R \cite{wang2024dust3r} generates point clouds from each image pair, and all resulting pointmaps are aligned using COLMAP \cite{schonberger2016structure} poses and their proposed alignment algorithm.
For EDGS, we initialize with pointmaps generated by our method and adopt the optimization strategy described in the EDGS paper.
Ours-align denotes our method using the initial alignment described in Section~\ref{section:initial_align}, skipping the geometric constraints and refinement optimization, and directly optimizing 3D Gaussian Splatting without densification.
Ours-full refers to our complete pipeline, including initial alignment, geometric constraints, refinement, and 3DGS optimization.

\begin{figure}[t]
    \centering
    \includegraphics[width=1.\linewidth]{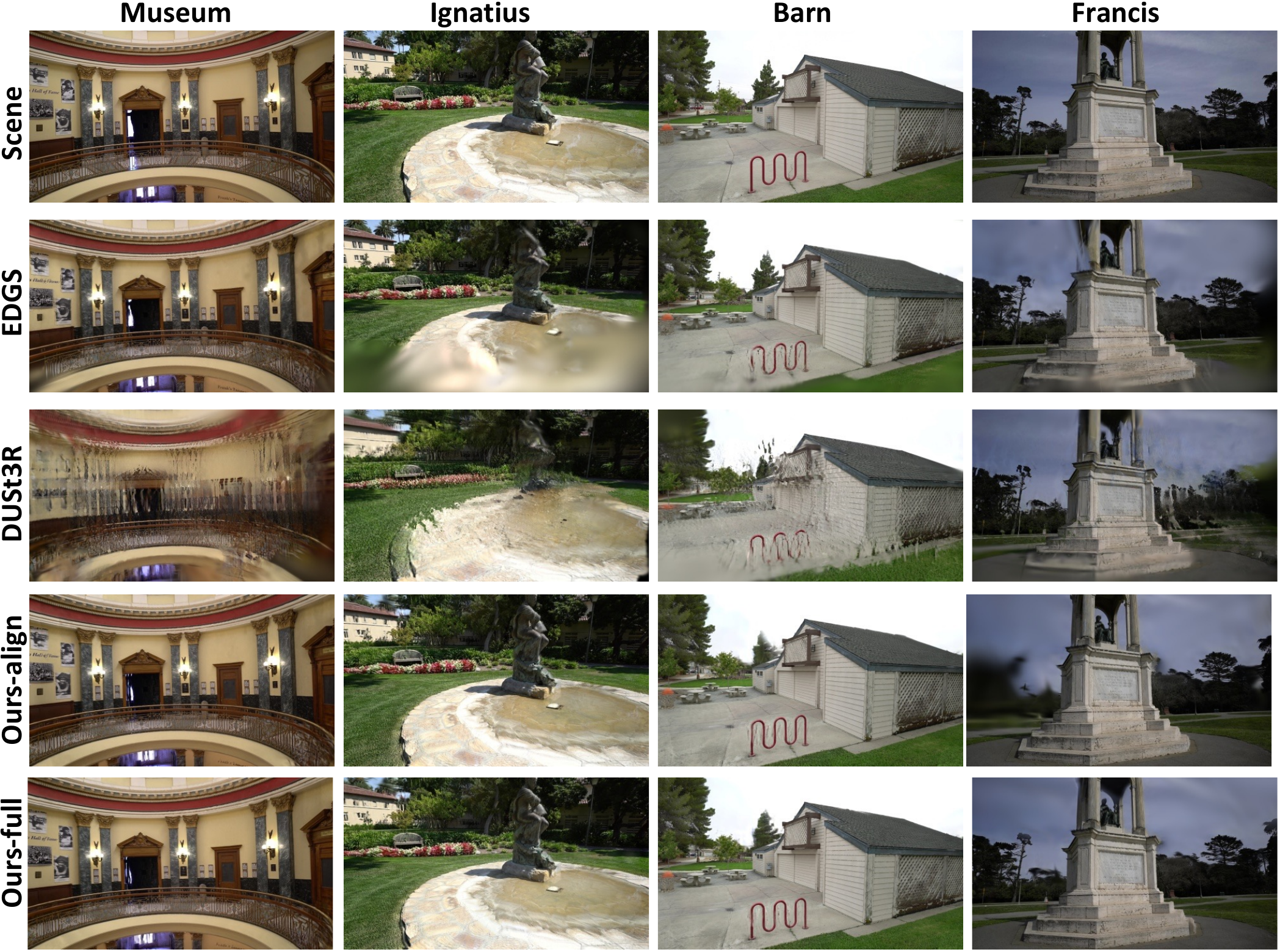}
    \caption{
    \textbf{Qualitative Comparison.} We compare 3D Gaussian Splatting (3DGS) \cite{kerbl20233d} initialized with different methods. Our aligned pointmap provides a better initialization, enabling higher rendering quality during optimization. 
    }
    \label{fig:gsplat}
\end{figure}

Table \ref{tab:gsplat_results} presents novel view synthesis results under sparse-view settings using 3, 6, and 12 training images.
Experiments are conducted with 200 and 1000 optimization steps, corresponding to the top and bottom rows, respectively.
Overall, 3DGS optimization using our monocular point map alignment consistently outperforms optimization using DUSt3R-aligned point maps. Notably, even with only 200 optimization steps, our method achieves high rendering quality due to the accurate initialization, which enables faster convergence without significant adjustments to the Gaussian positions.
In the ablation study, our full pipeline achieves slightly better rendering performance compared to using only the initial alignment step.
As shown in Fig. \ref{fig:gsplat}, we visualize the rendering results of different methods. The misalignment in DUSt3R \cite{wang2024dust3r} results in visible rendering artifacts.

\begin{table}
    \centering
    \begin{adjustbox}{width=0.99\linewidth,center}
    \begin{tabular}{l|c|cccccccccccccccccccccccc}
    \toprule
    \multirow{2}{*}{Method}&\multirow{2}{*}{Steps}&\multicolumn{3}{c}{3-view}&\multicolumn{3}{c}{6-view}&\multicolumn{3}{c}{12-view}\\
    & & {\scriptsize PSNR \(\uparrow\)} & {\scriptsize SSIM \(\uparrow\)} & {\scriptsize LPIPS \(\downarrow\)}
    & {\scriptsize PSNR \(\uparrow\)} & {\scriptsize SSIM \(\uparrow\)} & {\scriptsize LPIPS \(\downarrow\)}
    & {\scriptsize PSNR \(\uparrow\)} & {\scriptsize SSIM \(\uparrow\)} & {\scriptsize LPIPS \(\downarrow\)} \\
    \midrule
 DUSt3R \cite{wang2024dust3r}  & 200 &  \cellcolor{tabthird}13.77 &  \cellcolor{tabthird}0.386 & 0.558 &  \cellcolor{tabthird}14.35 & 0.406 & 0.560 &  \cellcolor{tabthird}14.40 & 0.426 & 0.567 \\
EDGS \cite{kotovenko2025edgs} & 200 & 8.79 & 0.293 &  \cellcolor{tabthird}0.548 & 9.98 &  \cellcolor{tabthird}0.412 &  \cellcolor{tabthird}0.466 & 10.65 &  \cellcolor{tabthird}0.475 &  \cellcolor{tabthird}0.425 \\
Ours-align   & 200    & \cellcolor{tabsecond}17.99 &  \cellcolor{tabfirst}0.608 &  \cellcolor{tabfirst}0.360 & \cellcolor{tabsecond}19.27 & \cellcolor{tabsecond}0.650 & \cellcolor{tabsecond}0.334 & \cellcolor{tabsecond}19.33 & \cellcolor{tabsecond}0.670 & \cellcolor{tabsecond}0.325 \\
Ours-full & 200  &  \cellcolor{tabfirst}18.48 & \cellcolor{tabsecond}0.544 & \cellcolor{tabsecond}0.366 &  \cellcolor{tabfirst}19.99 &  \cellcolor{tabfirst}0.662 &  \cellcolor{tabfirst}0.310 &  \cellcolor{tabfirst}20.12 &  \cellcolor{tabfirst}0.678 &  \cellcolor{tabfirst}0.304 \\ 
    \midrule
DUSt3R \cite{wang2024dust3r} & 1000 & 14.10 & 0.362 & 0.508 & 15.44 & 0.390 & 0.502 & 16.29 & 0.431 & 0.500 \\
EDGS \cite{kotovenko2025edgs} & 1000 &  \cellcolor{tabthird}18.53 &  \cellcolor{tabfirst}0.617 &  \cellcolor{tabthird}0.387 & \cellcolor{tabsecond}21.09 &  \cellcolor{tabfirst}0.697 &  \cellcolor{tabthird}0.304 & \cellcolor{tabsecond}22.32 &  \cellcolor{tabfirst}0.728 &  \cellcolor{tabthird}0.265 \\
Ours-align & 1000          & \cellcolor{tabsecond}19.03 & \cellcolor{tabsecond}0.523 &  \cellcolor{tabfirst}0.315 &  \cellcolor{tabthird}20.69 &  \cellcolor{tabthird}0.666 & \cellcolor{tabsecond}0.240 &  \cellcolor{tabthird}21.69 &  \cellcolor{tabthird}0.703 & \cellcolor{tabsecond}0.250 \\
Ours-full & 1000   &  \cellcolor{tabfirst}19.93 &  \cellcolor{tabthird}0.513 & \cellcolor{tabsecond}0.332 &  \cellcolor{tabfirst}21.52 & \cellcolor{tabsecond}0.682 &  \cellcolor{tabfirst}0.217 &  \cellcolor{tabfirst}22.39 & \cellcolor{tabsecond}0.715 &  \cellcolor{tabfirst}0.214 \\
    \bottomrule
    \end{tabular}
    \end{adjustbox}
    \caption{\textbf{Quantitative Comparison} on Tanks \& Temples dataset for novel-view synthesis. The top rows show results after 200 optimization steps, while the bottom rows show results after 1000 steps. Overall, our method outperforms other algorithms, particularly in terms of PSNR. The best, second-best, and third-best entries are marked in red, orange, and yellow, respectively. }
    \label{tab:gsplat_results}
\end{table}

\section{Conclusion, Limitations, and Future Work}
\label{sec:conclusion}
We present a novel framework for aligning monocular geometry across different views.
To address the inherent ambiguities in monocular geometry estimation, we introduce a cross-view affine alignment method based on feature matching.
This is followed by a joint graph optimization process that refines both point maps and surface normals, enhancing consistency between frames at the pixel level.
By leveraging the aligned point clouds, our method also improves rendering performance in sparse-view scenarios.
However, the current approach faces limitations in computational efficiency when processing multiple frames incrementally.
As future work, we plan to improve the parameterization process within the graph optimization and refine the overall pipeline, aiming to scale our method to broader multi-view reconstruction tasks.

\clearpage


\maketitlesupplementary
\setcounter{figure}{0}
\setcounter{table}{0}
\setcounter{section}{0}

\section{Performance Analysis}
\label{supp:sec:analysis}
\subsection{Runtime and Memory}
\begin{table}[h]
    \centering
    \begin{minipage}{1.0\linewidth}
        \centering
        \resizebox{\linewidth}{!}{
        \begin{tabular}{l|c|c|c|c|r}
        \toprule
        Type & Resolution & $\#$ Match & $k$ of kNN & GPU (L40S) & time \\
        \hline
        Default & $512\times377$ & 5000 & 4 & 838 MiB & 1.6 s \\
        Res. $\uparrow$ & \textbf{\boldmath $640\times471$} & 5000 & 4 & 1354 MiB & 2.5 s \\
        Matches $\uparrow$ & $512\times377$ & \textbf{10000} & 4 & 850 MiB & 1.6 s \\
        KNN k $\uparrow$ & $512\times377$ & 5000 & \textbf{16} & 1152 MiB & 2.0 s \\
        \bottomrule
        \end{tabular}
        }
        %
        \caption{Runtime and Memory. 
        The default represents our experimental setting, while the other types correspond to modified settings for analysis.}
        \label{table:rebuttal2}
    \end{minipage}
\end{table}
\noindent
Table \ref{table:rebuttal2} shows runtime and GPU memory usage with respect to resolution, the number of matching points, and the number of nearest neighbors $k$.

\subsection{Parameter Stability}
\begin{table}[h]
    \centering
    \begin{minipage}{1.\linewidth}
        \centering
        \resizebox{\linewidth}{!}{
        \begin{tabular}{l|ccccc|cc}
        \toprule
        Type $\downarrow$ & $\lambda_p$ & $\lambda_r$ & $\lambda_s$ & $\lambda_n$ & $k$ of kNN & rel$\downarrow$ & $\tau \uparrow$ \\
        \hline
        Default & 30 & 50 & 0.1 & 10 & 4 & 3.12 & 70.29 \\
        \hline
        $\lambda_p$ $\downarrow$& \textbf{15} & 50 & 0.1 & 10 & 4 & 3.13 & 70.23 \\
        $\lambda_p$ $\uparrow$& \textbf{45} & 50 & 0.1 & 10 & 4 & 3.13 & 70.20 \\
        \hline
        $\lambda_r$ $\downarrow$& 30 & \textbf{25} & 0.1 & 10 & 4 & 3.13 & 70.29 \\
        $\lambda_r$ $\uparrow$& 30 & \textbf{75} & 0.1 & 10 & 4 & 3.14 & 70.16 \\
        \hline
        $\lambda_s$ $\downarrow$& 30 & 50 & \textbf{0.05} & 10 & 4 & 3.13 & 70.21\\
        $\lambda_s$ $\uparrow$& 30 & 50 & \textbf{0.2} & 10 & 4 & 3.13 & 70.29 \\
        \hline
        $\lambda_n$ $\downarrow$& 30 & 50 & 0.1 & \textbf{5} & 4 & 3.13 & 70.22 \\
        $\lambda_n$ $\uparrow$& 30 & 50 & 0.1 & \textbf{20} & 4 & 3.13 & 70.21 \\
        \hline
        k $\uparrow$ & 30 & 50 & 0.1 & 10 & \textbf{16} & 3.12 & 70.36 \\
        \bottomrule
        \end{tabular}
        }
        \caption{Sensitivity to Parameter Settings. 
        The default represents our experimental setting, while the other types illustrate stability checks.}
        \label{table:rebuttal1}
    \end{minipage}
\end{table}
\noindent
Table \ref{table:rebuttal1} reports the performance in terms of Absolute Relative Error (rel) and Inlier Ratio ($\gamma$) with a threshold of 1.03 under varying parameter settings.
The results show overall consistent performance across different settings.
A larger number of nearest neighbors $k$ improves performance, but considering computational cost, we fix $k=4$ in our experiments.

\section{Additional Experimental Results}
\label{supp:sec:exp}

\subsection{3D Reconstruction}
\begin{figure*}[t]
    \centering
    \includegraphics[width=0.9\linewidth]{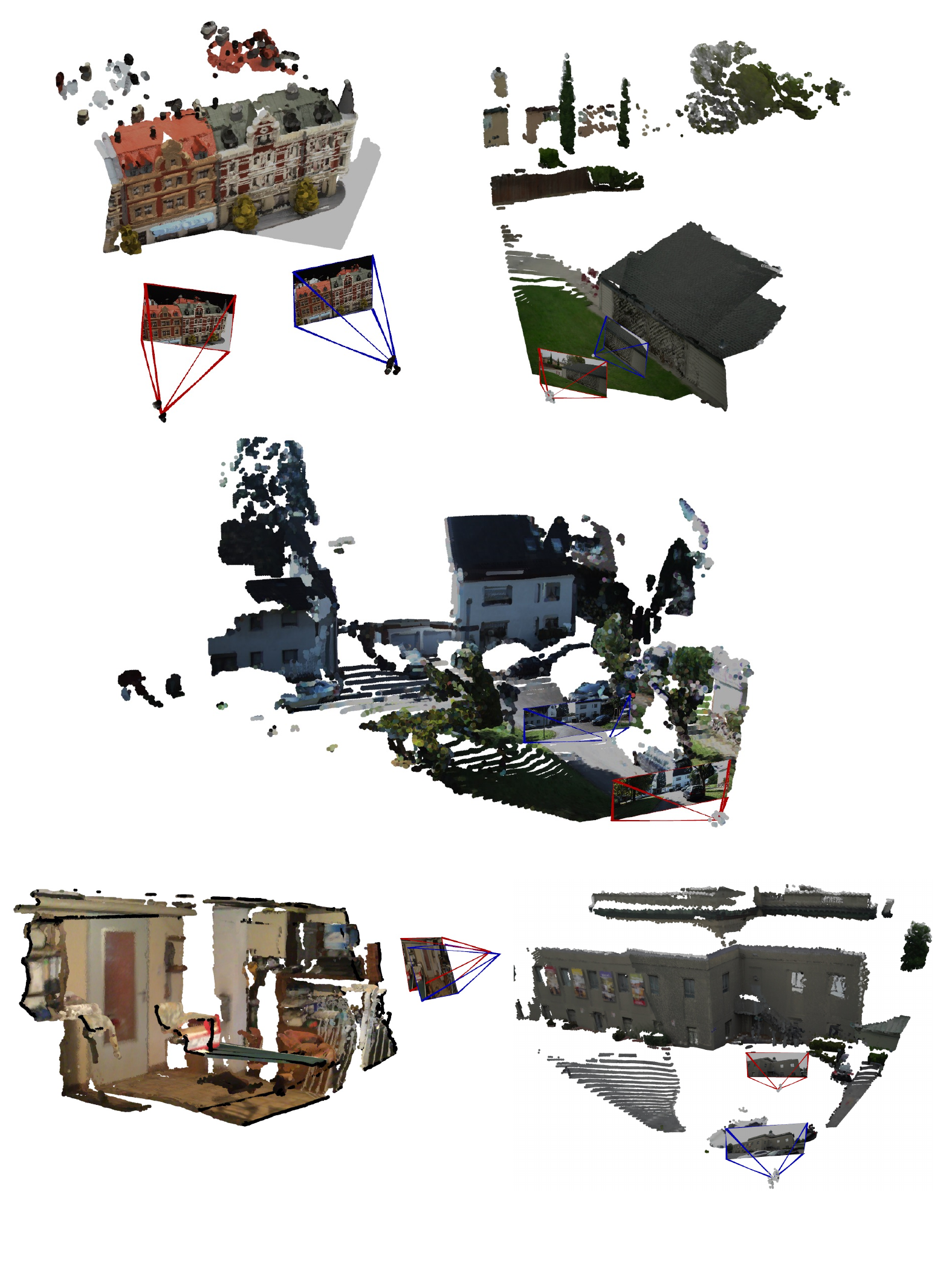}
    \caption{
    \textbf{Additional Qualitative Results of MoRe}
    }
    \label{fig:sup1}
\end{figure*}

\begin{figure*}[t]
    \centering
    \includegraphics[width=0.9\linewidth]{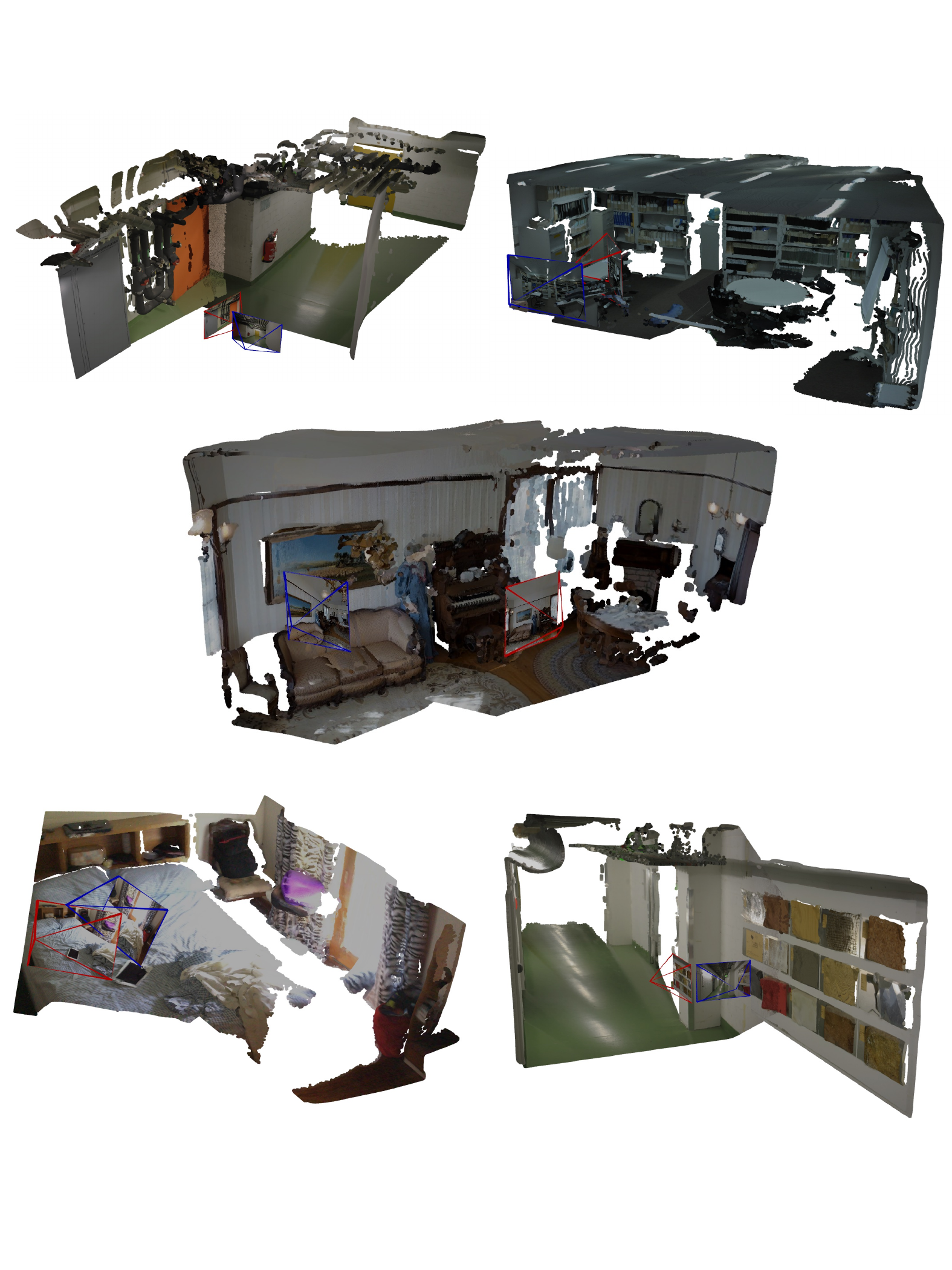}
    \caption{
    \textbf{Additional Qualitative Results of MoRe}
    }
    \label{fig:sup2}
\end{figure*}

We present additional qualitative results of our point alignment method.
Figure \ref{fig:sup1} shows the results for Case 1 (with given poses), and Figure \ref{fig:sup2} shows the results for Case 2 (without given poses).
We used the Tanks and Temples \cite{knapitsch2017tanks}, ETH3D \cite{schops2017multi}, ScanNet \cite{dai2017scannet}, Matterport3D \cite{chang2017matterport3d}, KITTI \cite{geiger2013vision}, and DTU \cite{aanaes2016large} datasets in this experiment.


\subsection{Novel View Synthesis}
To supplement Table 4 in the main paper, we present per-scene experimental results in terms of PSNR, SSIM, and LPIPS, as shown in Table \ref{tab:gsplat_results1}, \ref{tab:gsplat_results2} and \ref{tab:gsplat_results3}.

\begin{table*}
    \centering
    \begin{adjustbox}{width=0.99\linewidth,center}
    \begin{tabular}{l|c|cccccccccccccccccccccccc}
    \toprule
    \multirow{2}{*}{Method}&\multirow{2}{*}{Steps}&\multicolumn{3}{c}{Ballroom}&\multicolumn{3}{c}{Barn}&\multicolumn{3}{c}{Family}&\multicolumn{3}{c}{Francis}&\multicolumn{3}{c}{Horse}&\multicolumn{3}{c}{Ignatius}&\multicolumn{3}{c}{Museum}&\multicolumn{3}{c}{Mean} \\
    & & {\scriptsize PSNR \(\uparrow\)} & {\scriptsize SSIM \(\uparrow\)} & {\scriptsize LPIPS \(\downarrow\)}
    & {\scriptsize PSNR \(\uparrow\)} & {\scriptsize SSIM \(\uparrow\)} & {\scriptsize LPIPS \(\downarrow\)}
    & {\scriptsize PSNR \(\uparrow\)} & {\scriptsize SSIM \(\uparrow\)} & {\scriptsize LPIPS \(\downarrow\)}
    & {\scriptsize PSNR \(\uparrow\)} & {\scriptsize SSIM \(\uparrow\)} & {\scriptsize LPIPS \(\downarrow\)}
    & {\scriptsize PSNR \(\uparrow\)} & {\scriptsize SSIM \(\uparrow\)} & {\scriptsize LPIPS \(\downarrow\)}
    & {\scriptsize PSNR \(\uparrow\)} & {\scriptsize SSIM \(\uparrow\)} & {\scriptsize LPIPS \(\downarrow\)}
    & {\scriptsize PSNR \(\uparrow\)} & {\scriptsize SSIM \(\uparrow\)} & {\scriptsize LPIPS \(\downarrow\)}
    & {\scriptsize PSNR \(\uparrow\)} & {\scriptsize SSIM \(\uparrow\)} & {\scriptsize LPIPS \(\downarrow\)} \\
    \midrule
DUSt3R \cite{wang2024dust3r} & 200  &  \cellcolor{tabthird}10.65 &  \cellcolor{tabthird}0.202 & 0.592 &  \cellcolor{tabthird}15.60 & \cellcolor{tabsecond}0.497 &  \cellcolor{tabthird}0.490 &  \cellcolor{tabthird}11.54 &  \cellcolor{tabthird}0.351 & 0.615 &  \cellcolor{tabthird}16.85 &  \cellcolor{tabthird}0.520 & 0.508 &  \cellcolor{tabthird}12.92 &  \cellcolor{tabthird}0.512 &  \cellcolor{tabthird}0.523 &  \cellcolor{tabthird}15.16 & 0.306 & 0.587 &  \cellcolor{tabthird}13.65 &  \cellcolor{tabthird}0.312 & 0.592 &  \cellcolor{tabthird}13.77 &  \cellcolor{tabthird}0.386 & 0.558 \\
EDGS \cite{kotovenko2025edgs} & 200 & 9.98 & \cellcolor{tabsecond}0.338 & \cellcolor{tabsecond}0.533 & 5.99 & 0.214 & 0.610 & 7.93 & 0.326 &  \cellcolor{tabthird}0.558 & 10.93 & 0.224 &  \cellcolor{tabthird}0.505 & 3.71 & 0.144 & 0.624 & 10.59 &  \cellcolor{tabthird}0.368 &  \cellcolor{tabthird}0.521 & 12.37 & \cellcolor{tabsecond}0.440 &  \cellcolor{tabthird}0.486 & 8.79 & 0.293 &  \cellcolor{tabthird}0.548 \\
Ours-align  & 200                   &  \cellcolor{tabfirst}17.79 &  \cellcolor{tabfirst}0.561 &  \cellcolor{tabfirst}0.340 &  \cellcolor{tabfirst}17.78 &  \cellcolor{tabfirst}0.606 &  \cellcolor{tabfirst}0.364 & \cellcolor{tabsecond}16.46 & \cellcolor{tabsecond}0.644 & \cellcolor{tabsecond}0.340 &  \cellcolor{tabfirst}18.94 & \cellcolor{tabsecond}0.637 & \cellcolor{tabsecond}0.395 & \cellcolor{tabsecond}15.08 & \cellcolor{tabsecond}0.596 & \cellcolor{tabsecond}0.425 & \cellcolor{tabsecond}19.62 &  \cellcolor{tabfirst}0.523 & \cellcolor{tabsecond}0.402 &  \cellcolor{tabfirst}20.29 &  \cellcolor{tabfirst}0.689 & \cellcolor{tabsecond}0.254 & \cellcolor{tabsecond}17.99 &  \cellcolor{tabfirst}0.608 &  \cellcolor{tabfirst}0.360 \\
Ours-full & 200                     & \cellcolor{tabsecond}17.31 & \cellcolor{tabsecond}0.338 &  \cellcolor{tabthird}0.547 & \cellcolor{tabsecond}17.03 &  \cellcolor{tabthird}0.365 & \cellcolor{tabsecond}0.365 &  \cellcolor{tabfirst}18.70 &  \cellcolor{tabfirst}0.697 &  \cellcolor{tabfirst}0.291 & \cellcolor{tabsecond}18.73 &  \cellcolor{tabfirst}0.652 &  \cellcolor{tabfirst}0.350 &  \cellcolor{tabfirst}17.63 &  \cellcolor{tabfirst}0.670 &  \cellcolor{tabfirst}0.370 &  \cellcolor{tabfirst}19.69 & \cellcolor{tabsecond}0.396 &  \cellcolor{tabfirst}0.396 & \cellcolor{tabsecond}20.28 &  \cellcolor{tabfirst}0.689 &  \cellcolor{tabfirst}0.242 &  \cellcolor{tabfirst}18.48 & \cellcolor{tabsecond}0.544 & \cellcolor{tabsecond}0.366 \\ 
    \midrule
DUSt3R \cite{wang2024dust3r} & 1000  & 10.45 & 0.170 & \cellcolor{tabsecond}0.533 & 17.36 &  \cellcolor{tabthird}0.509 &  \cellcolor{tabthird}0.415 & 11.40 & 0.307 & 0.591 & 17.55 & 0.501 & 0.468 & 12.63 & 0.468 & 0.500 & 15.03 & 0.270 & 0.510 & 14.26 &  \cellcolor{tabthird}0.305 & 0.538 & 14.10 & 0.362 & 0.508 \\
EDGS \cite{kotovenko2025edgs} & 1000 &  \cellcolor{tabthird}17.03 &  \cellcolor{tabfirst}0.546 &  \cellcolor{tabfirst}0.383 &  \cellcolor{tabfirst}18.90 &  \cellcolor{tabfirst}0.628 & \cellcolor{tabsecond}0.383 & \cellcolor{tabsecond}19.31 & \cellcolor{tabsecond}0.670 &  \cellcolor{tabthird}0.404 &  \cellcolor{tabthird}20.86 &  \cellcolor{tabfirst}0.652 &  \cellcolor{tabthird}0.389 &  \cellcolor{tabthird}16.79 & \cellcolor{tabsecond}0.657 &  \cellcolor{tabthird}0.402 &  \cellcolor{tabthird}18.38 &  \cellcolor{tabfirst}0.545 &  \cellcolor{tabthird}0.418 &  \cellcolor{tabthird}18.46 & \cellcolor{tabsecond}0.625 &  \cellcolor{tabthird}0.333 &  \cellcolor{tabthird}18.53 &  \cellcolor{tabfirst}0.617 &  \cellcolor{tabthird}0.387 \\
Ours-align & 1000                    &  \cellcolor{tabfirst}18.32 &  \cellcolor{tabthird}0.245 & 0.568 &  \cellcolor{tabthird}18.09 & \cellcolor{tabsecond}0.579 &  \cellcolor{tabfirst}0.292 &  \cellcolor{tabthird}18.51 &  \cellcolor{tabthird}0.639 & \cellcolor{tabsecond}0.248 & \cellcolor{tabsecond}21.02 &  \cellcolor{tabthird}0.641 & \cellcolor{tabsecond}0.323 & \cellcolor{tabsecond}17.17 &  \cellcolor{tabthird}0.594 & \cellcolor{tabsecond}0.309 & \cellcolor{tabsecond}19.32 & \cellcolor{tabsecond}0.283 & \cellcolor{tabsecond}0.283 & \cellcolor{tabsecond}20.75 &  \cellcolor{tabfirst}0.682 & \cellcolor{tabsecond}0.184 & \cellcolor{tabsecond}19.03 & \cellcolor{tabsecond}0.523 &  \cellcolor{tabfirst}0.315 \\
Ours-full & 1000                     & \cellcolor{tabsecond}17.62 & \cellcolor{tabsecond}0.251 &  \cellcolor{tabthird}0.549 & \cellcolor{tabsecond}18.32 & 0.282 & 0.591 &  \cellcolor{tabfirst}21.10 &  \cellcolor{tabfirst}0.729 &  \cellcolor{tabfirst}0.210 &  \cellcolor{tabfirst}21.18 & \cellcolor{tabsecond}0.650 &  \cellcolor{tabfirst}0.287 &  \cellcolor{tabfirst}20.98 &  \cellcolor{tabfirst}0.721 &  \cellcolor{tabfirst}0.243 &  \cellcolor{tabfirst}19.47 &  \cellcolor{tabthird}0.273 &  \cellcolor{tabfirst}0.273 &  \cellcolor{tabfirst}20.82 &  \cellcolor{tabfirst}0.682 &  \cellcolor{tabfirst}0.171 &  \cellcolor{tabfirst}19.93 &  \cellcolor{tabthird}0.513 & \cellcolor{tabsecond}0.332 \\
    \bottomrule
    \end{tabular}
    \end{adjustbox}
    \caption{\textbf{Breakdown results} on Tanks \& Temples dataset for novel-view synthesis with \textbf{3 training views}. Red, orange, and yellow indicate the first, second, and third best performing algorithms for each metric.}
    \label{tab:gsplat_results1}
\end{table*}

\begin{table*}
    \centering
    \begin{adjustbox}{width=0.99\linewidth,center}
    \begin{tabular}{l|c|cccccccccccccccccccccccc}
    \toprule
    \multirow{2}{*}{Method}&\multirow{2}{*}{Steps}&\multicolumn{3}{c}{Ballroom}&\multicolumn{3}{c}{Barn}&\multicolumn{3}{c}{Family}&\multicolumn{3}{c}{Francis}&\multicolumn{3}{c}{Horse}&\multicolumn{3}{c}{Ignatius}&\multicolumn{3}{c}{Museum}&\multicolumn{3}{c}{Mean} \\
    & & {\scriptsize PSNR \(\uparrow\)} & {\scriptsize SSIM \(\uparrow\)} & {\scriptsize LPIPS \(\downarrow\)}
    & {\scriptsize PSNR \(\uparrow\)} & {\scriptsize SSIM \(\uparrow\)} & {\scriptsize LPIPS \(\downarrow\)}
    & {\scriptsize PSNR \(\uparrow\)} & {\scriptsize SSIM \(\uparrow\)} & {\scriptsize LPIPS \(\downarrow\)}
    & {\scriptsize PSNR \(\uparrow\)} & {\scriptsize SSIM \(\uparrow\)} & {\scriptsize LPIPS \(\downarrow\)}
    & {\scriptsize PSNR \(\uparrow\)} & {\scriptsize SSIM \(\uparrow\)} & {\scriptsize LPIPS \(\downarrow\)}
    & {\scriptsize PSNR \(\uparrow\)} & {\scriptsize SSIM \(\uparrow\)} & {\scriptsize LPIPS \(\downarrow\)}
    & {\scriptsize PSNR \(\uparrow\)} & {\scriptsize SSIM \(\uparrow\)} & {\scriptsize LPIPS \(\downarrow\)}
    & {\scriptsize PSNR \(\uparrow\)} & {\scriptsize SSIM \(\uparrow\)} & {\scriptsize LPIPS \(\downarrow\)} \\
    \midrule
DUSt3R \cite{wang2024dust3r} & 200  &  \cellcolor{tabthird}11.31 & 0.220 & 0.608 &  \cellcolor{tabthird}16.08 &  \cellcolor{tabthird}0.519 &  \cellcolor{tabthird}0.485 &  \cellcolor{tabthird}12.09 &  \cellcolor{tabthird}0.372 & 0.615 &  \cellcolor{tabthird}17.17 &  \cellcolor{tabthird}0.527 & 0.508 &  \cellcolor{tabthird}13.68 &  \cellcolor{tabthird}0.522 & 0.515 &  \cellcolor{tabthird}15.72 & 0.327 & 0.595 & 14.41 & 0.358 & 0.596 &  \cellcolor{tabthird}14.35 & 0.406 & 0.560 \\
EDGS \cite{kotovenko2025edgs} & 200 & 10.98 &  \cellcolor{tabthird}0.466 &  \cellcolor{tabthird}0.451 & 6.40 & 0.333 & 0.529 & 8.98 & \cellcolor{tabsecond}0.470 &  \cellcolor{tabthird}0.446 & 11.11 & 0.326 &  \cellcolor{tabthird}0.428 & 4.33 & 0.280 &  \cellcolor{tabthird}0.512 & 12.68 &  \cellcolor{tabthird}0.448 &  \cellcolor{tabthird}0.466 &  \cellcolor{tabthird}15.42 &  \cellcolor{tabthird}0.560 &  \cellcolor{tabthird}0.427 & 9.98 &  \cellcolor{tabthird}0.412 &  \cellcolor{tabthird}0.466 \\
Ours-align & 200                    & \cellcolor{tabsecond}19.02 & \cellcolor{tabsecond}0.611 & \cellcolor{tabsecond}0.319 &  \cellcolor{tabfirst}19.77 &  \cellcolor{tabfirst}0.681 & \cellcolor{tabsecond}0.310 & \cellcolor{tabsecond}18.14 &  \cellcolor{tabfirst}0.680 & \cellcolor{tabsecond}0.301 & \cellcolor{tabsecond}19.67 & \cellcolor{tabsecond}0.671 & \cellcolor{tabsecond}0.361 & \cellcolor{tabsecond}16.35 & \cellcolor{tabsecond}0.640 & \cellcolor{tabsecond}0.399 &  \cellcolor{tabfirst}20.32 &  \cellcolor{tabfirst}0.548 & \cellcolor{tabsecond}0.398 & \cellcolor{tabsecond}21.60 & \cellcolor{tabsecond}0.716 & \cellcolor{tabsecond}0.254 & \cellcolor{tabsecond}19.27 & \cellcolor{tabsecond}0.650 & \cellcolor{tabsecond}0.334 \\
Ours-full & 200                     &  \cellcolor{tabfirst}19.09 &  \cellcolor{tabfirst}0.644 &  \cellcolor{tabfirst}0.303 & \cellcolor{tabsecond}19.64 & \cellcolor{tabsecond}0.670 &  \cellcolor{tabfirst}0.307 &  \cellcolor{tabfirst}18.51 &  \cellcolor{tabfirst}0.680 &  \cellcolor{tabfirst}0.297 &  \cellcolor{tabfirst}23.09 &  \cellcolor{tabfirst}0.705 &  \cellcolor{tabfirst}0.275 &  \cellcolor{tabfirst}17.58 &  \cellcolor{tabfirst}0.669 &  \cellcolor{tabfirst}0.350 & \cellcolor{tabsecond}20.28 & \cellcolor{tabsecond}0.547 &  \cellcolor{tabfirst}0.395 &  \cellcolor{tabfirst}21.72 &  \cellcolor{tabfirst}0.722 &  \cellcolor{tabfirst}0.245 &  \cellcolor{tabfirst}19.99 &  \cellcolor{tabfirst}0.662 &  \cellcolor{tabfirst}0.310 \\
    \midrule
DUSt3R \cite{wang2024dust3r} & 1000  & 11.56 & 0.190 & 0.551 & 19.28 & 0.557 & 0.395 & 12.20 & 0.324 & 0.590 & 19.29 & 0.526 & 0.455 & 13.76 & 0.475 & 0.478 & 16.28 & 0.304 & 0.507 & 15.70 & 0.357 & 0.538 & 15.44 & 0.390 & 0.502 \\
EDGS \cite{kotovenko2025edgs} & 1000 &  \cellcolor{tabthird}18.84 & \cellcolor{tabsecond}0.655 &  \cellcolor{tabthird}0.297 &  \cellcolor{tabthird}20.95 &  \cellcolor{tabfirst}0.682 &  \cellcolor{tabthird}0.314 &  \cellcolor{tabfirst}22.16 &  \cellcolor{tabfirst}0.749 &  \cellcolor{tabthird}0.304 & \cellcolor{tabsecond}23.73 &  \cellcolor{tabfirst}0.737 & \cellcolor{tabsecond}0.303 &  \cellcolor{tabfirst}20.46 &  \cellcolor{tabfirst}0.747 &  \cellcolor{tabthird}0.301 &  \cellcolor{tabfirst}20.82 &  \cellcolor{tabfirst}0.604 &  \cellcolor{tabthird}0.354 &  \cellcolor{tabthird}20.69 &  \cellcolor{tabthird}0.708 &  \cellcolor{tabthird}0.257 & \cellcolor{tabsecond}21.09 &  \cellcolor{tabfirst}0.697 &  \cellcolor{tabthird}0.304 \\
Ours-align & 1000                    & \cellcolor{tabsecond}20.17 &  \cellcolor{tabthird}0.640 & \cellcolor{tabsecond}0.218 & \cellcolor{tabsecond}21.17 & \cellcolor{tabsecond}0.672 & \cellcolor{tabsecond}0.229 & \cellcolor{tabsecond}20.19 &  \cellcolor{tabthird}0.690 &  \cellcolor{tabfirst}0.212 &  \cellcolor{tabthird}20.29 &  \cellcolor{tabthird}0.697 &  \cellcolor{tabthird}0.318 &  \cellcolor{tabthird}19.81 &  \cellcolor{tabthird}0.682 & \cellcolor{tabsecond}0.269 &  \cellcolor{tabthird}20.54 &  \cellcolor{tabthird}0.545 & \cellcolor{tabsecond}0.263 & \cellcolor{tabsecond}22.63 & \cellcolor{tabsecond}0.734 & \cellcolor{tabsecond}0.170 &  \cellcolor{tabthird}20.69 &  \cellcolor{tabthird}0.666 & \cellcolor{tabsecond}0.240 \\
Ours-full & 1000                     &  \cellcolor{tabfirst}20.69 &  \cellcolor{tabfirst}0.689 &  \cellcolor{tabfirst}0.202 &  \cellcolor{tabfirst}21.60 &  \cellcolor{tabthird}0.668 &  \cellcolor{tabfirst}0.220 &  \cellcolor{tabthird}20.03 & \cellcolor{tabsecond}0.692 & \cellcolor{tabsecond}0.213 &  \cellcolor{tabfirst}24.27 & \cellcolor{tabsecond}0.731 &  \cellcolor{tabfirst}0.234 & \cellcolor{tabsecond}20.31 & \cellcolor{tabsecond}0.692 &  \cellcolor{tabfirst}0.238 & \cellcolor{tabsecond}20.59 & \cellcolor{tabsecond}0.551 &  \cellcolor{tabfirst}0.258 &  \cellcolor{tabfirst}23.15 &  \cellcolor{tabfirst}0.750 &  \cellcolor{tabfirst}0.157 &  \cellcolor{tabfirst}21.52 & \cellcolor{tabsecond}0.682 &  \cellcolor{tabfirst}0.217 \\
    \bottomrule
    \end{tabular}
    \end{adjustbox}
    \caption{\textbf{Breakdown results} on Tanks \& Temples dataset for novel-view synthesis with \textbf{6 training views}. Red, orange, and yellow indicate the first, second, and third best performing algorithms for each metric.}
    \label{tab:gsplat_results2}
\end{table*}

\begin{table*}
    \centering
    \begin{adjustbox}{width=0.99\linewidth,center}
    \begin{tabular}{l|c|cccccccccccccccccccccccc}
    \toprule
    \multirow{2}{*}{Method}&\multirow{2}{*}{Steps}&\multicolumn{3}{c}{Ballroom}&\multicolumn{3}{c}{Barn}&\multicolumn{3}{c}{Family}&\multicolumn{3}{c}{Francis}&\multicolumn{3}{c}{Horse}&\multicolumn{3}{c}{Ignatius}&\multicolumn{3}{c}{Museum}&\multicolumn{3}{c}{Mean} \\
    & & {\scriptsize PSNR \(\uparrow\)} & {\scriptsize SSIM \(\uparrow\)} & {\scriptsize LPIPS \(\downarrow\)}
    & {\scriptsize PSNR \(\uparrow\)} & {\scriptsize SSIM \(\uparrow\)} & {\scriptsize LPIPS \(\downarrow\)}
    & {\scriptsize PSNR \(\uparrow\)} & {\scriptsize SSIM \(\uparrow\)} & {\scriptsize LPIPS \(\downarrow\)}
    & {\scriptsize PSNR \(\uparrow\)} & {\scriptsize SSIM \(\uparrow\)} & {\scriptsize LPIPS \(\downarrow\)}
    & {\scriptsize PSNR \(\uparrow\)} & {\scriptsize SSIM \(\uparrow\)} & {\scriptsize LPIPS \(\downarrow\)}
    & {\scriptsize PSNR \(\uparrow\)} & {\scriptsize SSIM \(\uparrow\)} & {\scriptsize LPIPS \(\downarrow\)}
    & {\scriptsize PSNR \(\uparrow\)} & {\scriptsize SSIM \(\uparrow\)} & {\scriptsize LPIPS \(\downarrow\)}
    & {\scriptsize PSNR \(\uparrow\)} & {\scriptsize SSIM \(\uparrow\)} & {\scriptsize LPIPS \(\downarrow\)} \\
    \midrule
DUSt3R \cite{wang2024dust3r} & 200  &  \cellcolor{tabthird}11.68 & 0.234 & 0.620 &  \cellcolor{tabthird}15.64 &  \cellcolor{tabthird}0.525 & 0.502 &  \cellcolor{tabthird}12.33 & 0.394 & 0.621 &  \cellcolor{tabthird}16.87 &  \cellcolor{tabthird}0.560 & 0.493 &  \cellcolor{tabthird}14.04 &  \cellcolor{tabthird}0.542 & 0.513 &  \cellcolor{tabthird}15.59 & 0.336 & 0.612 & 14.67 & 0.387 & 0.608 &  \cellcolor{tabthird}14.40 & 0.426 & 0.567 \\
EDGS \cite{kotovenko2025edgs} & 200 & 11.58 &  \cellcolor{tabthird}0.519 &  \cellcolor{tabthird}0.412 & 7.00 & 0.423 &  \cellcolor{tabthird}0.479 & 9.60 &  \cellcolor{tabthird}0.556 &  \cellcolor{tabthird}0.385 & 11.60 & 0.369 &  \cellcolor{tabthird}0.423 & 4.74 & 0.349 &  \cellcolor{tabthird}0.468 & 12.93 &  \cellcolor{tabthird}0.482 &  \cellcolor{tabthird}0.442 &  \cellcolor{tabthird}17.11 &  \cellcolor{tabthird}0.626 &  \cellcolor{tabthird}0.368 & 10.65 &  \cellcolor{tabthird}0.475 &  \cellcolor{tabthird}0.425 \\
Ours-align & 200                    &  \cellcolor{tabfirst}19.35 &  \cellcolor{tabfirst}0.663 &  \cellcolor{tabfirst}0.286 & \cellcolor{tabsecond}20.06 &  \cellcolor{tabfirst}0.702 & \cellcolor{tabsecond}0.295 & \cellcolor{tabsecond}18.12 & \cellcolor{tabsecond}0.702 & \cellcolor{tabsecond}0.299 & \cellcolor{tabsecond}19.39 & \cellcolor{tabsecond}0.670 & \cellcolor{tabsecond}0.365 & \cellcolor{tabsecond}15.98 & \cellcolor{tabsecond}0.650 & \cellcolor{tabsecond}0.394 &  \cellcolor{tabfirst}20.31 &  \cellcolor{tabfirst}0.564 & \cellcolor{tabsecond}0.398 &  \cellcolor{tabfirst}22.10 &  \cellcolor{tabfirst}0.739 & \cellcolor{tabsecond}0.240 & \cellcolor{tabsecond}19.33 & \cellcolor{tabsecond}0.670 & \cellcolor{tabsecond}0.325 \\
Ours-full & 200                     & \cellcolor{tabsecond}19.07 & \cellcolor{tabsecond}0.655 & \cellcolor{tabsecond}0.293 &  \cellcolor{tabfirst}20.11 & \cellcolor{tabsecond}0.698 &  \cellcolor{tabfirst}0.292 &  \cellcolor{tabfirst}21.14 &  \cellcolor{tabfirst}0.736 &  \cellcolor{tabfirst}0.212 &  \cellcolor{tabfirst}20.60 &  \cellcolor{tabfirst}0.700 &  \cellcolor{tabfirst}0.330 &  \cellcolor{tabfirst}17.64 &  \cellcolor{tabfirst}0.670 &  \cellcolor{tabfirst}0.370 & \cellcolor{tabsecond}20.20 & \cellcolor{tabsecond}0.554 &  \cellcolor{tabfirst}0.396 & \cellcolor{tabsecond}22.08 & \cellcolor{tabsecond}0.730 &  \cellcolor{tabfirst}0.234 &  \cellcolor{tabfirst}20.12 &  \cellcolor{tabfirst}0.678 &  \cellcolor{tabfirst}0.304 \\
    \midrule
DUSt3R \cite{wang2024dust3r} & 1000  & 11.99 & 0.213 & 0.578 & 19.80 &  \cellcolor{tabthird}0.595 & 0.395 & 13.13 & 0.355 & 0.592 & 21.25 & 0.622 & 0.392 & 14.79 & 0.501 & 0.469 & 16.69 & 0.331 & 0.526 & 16.36 & 0.402 & 0.550 & 16.29 & 0.431 & 0.500 \\
EDGS \cite{kotovenko2025edgs} & 1000 &  \cellcolor{tabthird}19.25 &  \cellcolor{tabthird}0.687 &  \cellcolor{tabthird}0.264 & \cellcolor{tabsecond}22.98 &  \cellcolor{tabfirst}0.734 & \cellcolor{tabsecond}0.260 &  \cellcolor{tabfirst}23.60 &  \cellcolor{tabfirst}0.794 & \cellcolor{tabsecond}0.253 & \cellcolor{tabsecond}24.61 &  \cellcolor{tabfirst}0.762 & \cellcolor{tabsecond}0.277 &  \cellcolor{tabfirst}21.51 &  \cellcolor{tabfirst}0.783 & \cellcolor{tabsecond}0.249 &  \cellcolor{tabfirst}21.85 &  \cellcolor{tabfirst}0.632 &  \cellcolor{tabthird}0.335 &  \cellcolor{tabthird}22.47 &  \cellcolor{tabthird}0.703 &  \cellcolor{tabthird}0.218 & \cellcolor{tabsecond}22.32 &  \cellcolor{tabfirst}0.728 &  \cellcolor{tabthird}0.265 \\
Ours-align & 1000                    &  \cellcolor{tabfirst}21.92 &  \cellcolor{tabfirst}0.728 &  \cellcolor{tabfirst}0.186 &  \cellcolor{tabthird}22.68 & \cellcolor{tabsecond}0.698 &  \cellcolor{tabthird}0.292 &  \cellcolor{tabthird}18.70 &  \cellcolor{tabthird}0.697 &  \cellcolor{tabthird}0.291 &  \cellcolor{tabthird}23.01 &  \cellcolor{tabthird}0.720 &  \cellcolor{tabthird}0.280 &  \cellcolor{tabthird}20.36 &  \cellcolor{tabthird}0.705 &  \cellcolor{tabthird}0.271 & \cellcolor{tabsecond}21.24 & \cellcolor{tabsecond}0.593 & \cellcolor{tabsecond}0.273 &  \cellcolor{tabfirst}23.93 &  \cellcolor{tabfirst}0.778 & \cellcolor{tabsecond}0.159 &  \cellcolor{tabthird}21.69 &  \cellcolor{tabthird}0.703 & \cellcolor{tabsecond}0.250 \\
Ours-full & 1000                     & \cellcolor{tabsecond}21.42 & \cellcolor{tabsecond}0.713 & \cellcolor{tabsecond}0.197 &  \cellcolor{tabfirst}23.28 &  \cellcolor{tabfirst}0.734 &  \cellcolor{tabfirst}0.190 & \cellcolor{tabsecond}21.11 & \cellcolor{tabsecond}0.729 &  \cellcolor{tabfirst}0.210 &  \cellcolor{tabfirst}25.00 & \cellcolor{tabsecond}0.753 &  \cellcolor{tabfirst}0.232 & \cellcolor{tabsecond}20.98 & \cellcolor{tabsecond}0.721 &  \cellcolor{tabfirst}0.243 &  \cellcolor{tabthird}21.20 &  \cellcolor{tabthird}0.590 &  \cellcolor{tabfirst}0.270 & \cellcolor{tabsecond}23.75 & \cellcolor{tabsecond}0.767 &  \cellcolor{tabfirst}0.155 &  \cellcolor{tabfirst}22.39 & \cellcolor{tabsecond}0.715 &  \cellcolor{tabfirst}0.214 \\ 
    \bottomrule
    \end{tabular}
    \end{adjustbox}
    \caption{\textbf{Breakdown results} on Tanks \& Temples dataset for novel-view synthesis with \textbf{12 training views}. Red, orange, and yellow indicate the first, second, and third best performing algorithms for each metric.}
    \label{tab:gsplat_results3}
\end{table*}

\clearpage
\newpage
{
    \small
    \bibliographystyle{ieeenat_fullname}
    \bibliography{main}
}

\end{document}